\documentclass{article}

\usepackage{arxiv}
\usepackage{times}
\usepackage{xcolor}
\usepackage{soul}
\usepackage[utf8]{inputenc}
\usepackage[small]{caption}
\usepackage{amsmath}
\usepackage[]{multirow}

\usepackage{yfonts}
\usepackage{eufrak}
\usepackage{dsfont}
\usepackage[mathcal]{euscript}
\usepackage{txfonts}
\usepackage{hyperref}
\usepackage{graphicx}

\def \I {\varmathbb I}

\def \E {\varmathbb E}
\usepackage[utf8]{inputenc} 
\usepackage[T1]{fontenc}    
\usepackage{hyperref}       
\usepackage{url}            
\usepackage{booktabs}       
\usepackage{amsfonts}       
\usepackage{nicefrac}       
\usepackage{microtype}      
\usepackage{lipsum}
\title{Explainability in Human-Agent Systems}

\author{
Avi Rosenfeld \\
              Department of Computer Science of the
              Jerusalem College of Technology\\
              Jerusalem, Israel 91160\\
  \texttt{rosenfa@jct.ac.il} \\
   \And
 Ariella Richardson \\
              Department of Industrial Engineering  of the
              Jerusalem College of Technology\\
              Jerusalem, Israel 91160\\
  \texttt{richards@jct.ac.il} \\
}

\begin{document}
\maketitle

\begin{abstract}

This paper presents a taxonomy of explainability in Human-Agent Systems. We consider fundamental questions about the \emph{Why, Who, What, When and How} of explainability. First, we define explainability, and its relationship to the related terms of interpretability, transparency, explicitness, and faithfulness. These definitions allow us to answer \emph{why} explainability is needed in the system, \emph{whom} it is geared to and \emph{what} explanations can be generated to meet this need. We then consider \emph{when} the user should be presented with this information. Last, we consider \emph{how} objective and subjective measures can be used to evaluate the entire system. This last question is the most encompassing as it will need to evaluate all other issues regarding explainability.
\end{abstract}


\section{Introduction}
\label{Intro}

As the field of Artificial Intelligence matures and becomes ubiquitous, there is a growing emergence of systems where people and agents work together.  These systems, often called Human-Agent Systems or Human-Agent Cooperatives, have moved from theory to reality in the many forms, including digital personal assistants, recommendation systems, training and tutoring systems, service robots, chat bots, planning systems and self-driving cars \cite{amir2013plan,azaria2015strategic,barrett2017making,biran2017explanation,XAIP2017,jennings2014human,kleinerman2018providing,langley2017explainable,richardson2008coach,rosenfeld2017intelligent,rosenfeld2012learning,rosenfeld2015learning,salem2015would,sheh2017did,sierhuis2003human,traum2003negotiation,vanlehn2011relative,xiao2007commerce}. One key question surrounding these systems is the type and quality of the information that must be shared between the agents and the human-users during their interactions.  

This paper focuses on one aspect of this human-agent interaction — the internal level of explainability that agents using machine learning must have regarding the decisions they make. The overall goal of this paper is to provide an extensive study of this issue in Human-Agent Systems. Towards this goal, our first step is to formally and clearly define explainability in Section \ref{sec:definitions}, as well as the concepts of interpretability, transparency, explicitness, and faithfulness that make a system explainable. Through using these definitions, we provide a clear taxonomy regarding the \emph{Why, Who, What, When, and How} about explainability and stress the relationship of interpretability, transparency, explicitness, and faithfulness to each of these issues.  

Overall, we believe that the solutions presented to all of these issues need to be considered in tandem as they are intertwined. The type of explainability needed directly depends on the motivation for the type of human-agent system being implemented and thus directly stems from the first question about the overall reason, or reasons, for why the system must be explainable. Assuming that the system is human-centric, as is the case in recommendation \cite{kleinerman2018providing,xiao2007commerce}, training \cite{traum2003negotiation}, and tutoring systems \cite{amir2013plan,vanlehn2011relative}, then the information will likely need to persuade the person to choose a certain action, for example through arguments about the agent's decision   \cite{rosenfeld2016providing}, its  policy \cite{rosenfeld2017intelligent} or presentation \cite{azaria2014agent} . If the system is agent-centric, such as in knowledge discovery or self-driving cars, the agent might need to provide information about its decision to help convince the human participant of the correctness of their solution, aiding in the adoption of these agent based technologies \cite{Ribeiro2016}. In both cases, the information the agent provides should build trust to ensure its decisions are accepted \cite{abs-1806-00069,Guidotti2018,jennings2014human,lee2004trust}. Furthermore, these explanations might be necessary for legal considerations \cite{doshi2017towards,vlek2016method}. In all cases we need to consider and then evaluate \emph{how} these explanations were generated, presented, and if their level of detail correctly matches the system's need, something we address in Section \ref{How}.  

This paper is structured as follows. First, in Section \ref{sec:definitions}, we provide definitions for the terms of explainability, interpretability, transparency, fairness, explicitness and faithfulness and discuss the relationship between these terms. Based on these definitions, in Section \ref{Why} we present a taxonomy of three possibilities for \emph{why} explainability might be needed, ranging from not helpful, beneficial and critical. In Section \ref{Who}, we suggest three possible targets for \emph{who} the explanation is geared for: ``regular users", ``expert users", or entities external to the users of the system.  In Section \ref{What}, we address \emph{what} mechanism is used to create explanations. We consider six possibilities: directly from the machine learning algorithm, using feature selection and analysis, through a tool separate from the learning algorithm to model all definitions, a tool to explain a specific outcome, visualization tools and prototype analysis. In Section \ref{when} we address \emph{when} the generated explanations should be presented: before, after and/or during the task execution. In Section \ref{How} we introduce a general framework to evaluate explanations. Section \ref{discuss} includes a discussion about the taxonomy presented in the paper, including a table summarizing previous works and how they relate. Section \ref{conclusion} concludes.

\section{Definitions of Explainable Systems and Related Terms}
\label{sec:definitions}
 Several works have focused on the definitions of a system's explainability and also the related definitions of interpretability, transparency, fairness, explicitness and faithfulness.  As we demonstrate in this section, of all of these terms, we believe that the objective of making a system explainable is the most central and important for three reasons. Chronologically, this term was introduced first and thus has largest research history. Second, and possibly due to the first factor, this is the most general term. As we explain in this section, a system's level of explainability is created through the interpretations that the agent provides. These interpretable elements can be transparent, fair, explicit, and/or faithful. Last, and most importantly, this term connotes the key objective for the system: facilitating the human user's understanding of the agent's logic.  

\subsection{Theoretical Foundations for Explainability}
It has been noted that a thorough study of the term explanation would need to start with Aristotle as since his time it has been noted that explanations and causal reasoning are intrinsically intertwined \cite{hoffman2017explaining}. Specific to computer systems, as early as 1982, expert systems such as MYCIN and NEOMYCIN were developed for encoding the logical process within complex systems \cite{clancey1983epistemology,clancey1982neomycin}.  The objective of these systems, as is still the case, was to provide a set of clear explanations for a complex process.  However, no clear definitions for the nature of what constituted an  explanation  was provided.

Work by Gregor and Benbasat in 1999 defined the nature of explainability within ``intelligent" or ``knowledge-based" systems as a ``declaration of the meaning of words spoken, actions, motives, etc., with a view to adjusting a misunderstanding or
reconciling differences" \cite{gregor1999explanations}.  As they point out in their paper, this definition assumes that the explanation is provided by the provider of the information, in our case the intelligent agent, and that the explanation is geared to resolve some type of misunderstanding or disagreement. This definition is in line with other work that assumed that explanations were needed to help understand a system malfunction, an anomaly or to resolve conflict between the system and the user \cite{gilbert1989explanation,ortony1987surprisingness,schank1986explanation}.  Given this definition, it is not surprising that the first agent explanations were basic reasoning traces that assume the user will understand the technical information provided,  without taking a user other than the system designer into account. As these explanations are not typically processed beyond the raw logic of the system, they are referred to as ``na\"ive explanations" by previous work \cite{sormo2004explanation}. In our opinion, explainability of this type is more appropriate for system debugging than for other uses.

Possibly more generally, the Philosophy of Science community also provided several definitions of explainability. Most similar to the previous definition, work by Schank \cite{schank1986explanation} specifies that explanations address anomalies where a person is faced with a situation that does not fit her internalized model of the world. This type of definition can be thought of as goal-based, as the goal of the explanation is to address a specific need (e.g. disharmony within a user's internalized model) \cite{sormo2004explanation}.  Thus, explanations focus on an operational goal of addressing why the system isn't functioning as expected.

A second theory by van Fraassen \cite{van198511} claims that an explanation is always an answer to an implicit or explicit why-question comparing two or more possibilities. As such, an explanation provides information about why possibility $S_0$ was chosen and not options $S_1 \dots S_n$ \cite{sormo2004explanation,van198511}.  This definition suggests a minimum criteria any explanation must fulfill, namely that it facilitates a user choosing a specific option $S_0$, as well as a framework for understanding explanations as answers to why-questions contrasting two or more states \cite{sormo2004explanation}. One limitation of this approach is that the provided explanation has no use beyond helping the user understand why possibility $S_0$ was preferable relative to other possibilities. 

Most generally, a third theory by Achinstein \cite{achinstein1983nature} focuses on explanations as a process of communication between people. Here, the goal of an explanation is to provide the knowledge a recipient requests from a designated sender. Accordingly, this theory does not necessarily  require a complete explanation if the system's user does not require it. Consider a previously described example \cite{sormo2005explanation} that a neural network is trained to compare two pictures of a certain type and can give a similarity measure, e.g. from 0 to 1, and most people cannot understand how it came up with this score. Presenting the pictures to the user so she can validate the similarity for herself can itself serve as an explanation. As the very definition of a proper explanation is dependent on the interaction between the sender and the receiver, such an explanation is sufficient. Similarly, explanations can be motivated by many situations and not exclusively van Fraassen's why-questions. Conversely, a proper definition can and should be limited only to the information needed to address the receiver's request. 

\subsection{The Need for Precisely Defining Explainability in Human-Agent Systems}
Recently, questions have arose as to the definition of explainability of machine learning and agent systems. An explosive growth of interest has been registered within various research communities as is evident by workshops on: Explanation-aware Computing (ExaCt), Fairness, Accountability, and Transparency (FAT-ML), Workshop on Human Interpretability in Machine Learning (WHI), Interpretable ML for Complex Systems, Workshop on Explainable AI, Human-Centred Machine Learning, and Explainable Smart Systems \cite{CHI2018}.  However, no consensus exists about the meaning of various terms related to explainability including interpretability, transparency, explicitness, and faithfulness. It has been pointed out that the Oxford English dictionary does not have a definition for the term ``explainable"  \cite{DoranSB17}. One definition for an explanation that has been suggested as a, ``statement or account that makes something clear; a reason or justification given for an action or belief" is not always true for systems that claim to be explainable \cite{DoranSB17}.  Thus, providing an accepted and unified definition of explainability and other related terms is of great importance.

Part of the confusion is likely complicated by the fact that the terms, ``explainability, interpretability and transparency" are often used synonymously while other researchers implicitly define these terms differently \cite{DoranSB17,doshi2017towards,abs-1806-00069,Guidotti2018,Lipton16a,samek2017explainable}.  Artificial intelligence researchers tend to use the term Explainable AI (XAI) \cite{CHI2018,gunning2017explainable}, and focus on how explainable an artificial intelligence (XAI) system is without necessarily directly addressing the machine learning algorithms.  For example, work on explainable planning, which they coin XAIP, takes a system view of planning without considering any machine learning algorithms. They distance themselves from machine learning and deep learning systems which they claim are still far from being explainable \cite{XAIP2017}. 

In contrast, the machine learning community often focuses on the ``interpretability" of a machine learning system by focusing on how a machine learning algorithm makes its decisions and how interpretations can be derived either directly or secondarily from the machine learning component \cite{doshi2017towards,letham2015interpretable,rudin2014algorithms,vellido2012making,wang2017bayesian}. However, this term is equally poorly defined. In fact, one paper has gone so far as to recently write that, ``at present, interpretability has no formal technical meaning" and that, ``the term interpretability holds no agreed upon meaning, and yet machine
learning conferences frequently publish papers which wield the term in a quasimathematical way” \cite{Lipton16a}. In these papers, there is no syntactical technical difference between interpretable and explainable systems, as both terms refer to aspects of providing information to a human actor about the agent's decision-making process. Previous work generally defined interpretability as the ability to explain or present the decisions of a machine learning system using understandable terms  \cite{doshi2017towards}. More technically, Montanavon et al. propose that ``an interpretation is the mapping of an abstract concept (e.g. a predicted class) into a domain that the human can make sense of" which in turn forms explanations \cite{post}. Similarly, Doran et al. define interpretability as ``a system where a user cannot only see, but also study and understand how inputs are mathematically mapped to outputs."  To them, the opposite of interpretable systems are ``opaque" or ``black box" systems which yield no insight about the mapping between a decision and the inputs that yielded that decision \cite{DoranSB17}. 

Within the Machine Learning / Agent community, transparency has been informally defined to be the opposite of opacity or ``blackbox-ness" \cite{Lipton16a}. In order to clarify the difference between interpretability and transparency, we build upon the definition of transparency as an explanation about how the system reached its conclusion \cite{sormo2005explanation}.  More formally, transparency has been defined as a decision model where the decision-making process can be directly understood without any additional information \cite{Guidotti2018}. It is generally accepted that certain decision models are inherently transparent and others are not. For example, decision trees, and especially relatively small decision trees, are transparent, while deep neural networks cannot be understood without the aid of a explanation tool outside that of the decision process \cite{Guidotti2018}. We consider this difference in the next section and again in Section \ref{What}.

\label{inter_subsection}

\subsection{Formal Definitions for Explainability, Interpretability and Transparency in Human-Agent Systems}
\label{definitions}
This paper's first contribution is a clear definition for explainability and for the related terms: interpretability and transparency. In defining these terms we also define how explicitness and faithfulness are used within the context of Human-Agent Systems.  A summary of these definitions is found in Table \ref{table2}. 

In defining these terms, we focus on the features and records that are used as training input in the system, the supervised targets that need to be identified, and the machine learning algorithm used by the agent. We define $L$ as the machine learning algorithm that is created from a set of training records, $R$.
Each record $r \in R$ contains values for a tuple of ordered features, $F$. 
Each feature is defined as $f \in F$.  Thus, the entire training set consists of $R \times F$.  For example: Assume that the features are: $f1 = age$ (years), $f2 = height$ (cm), $f3 = weight$ (kg), so $F = \{age, height, weight\}$. A possible record  $r \in R$  might be $r=\{35,160,70\}$. While this model naturally lends itself to tabular data, it can as easily be applied to other forms of input such as texts, whereby $f$ are strings, or images whereby $f$ are pixels. The objective of $L$ is to properly fit $R \times F$ with regard to the labeled targets $t \in T$.

\begin{table}[]
\begin{tabular}{|l|c|l|}
\hline
Term                                                                     & \begin{tabular}[c]{@{}c@{}}Notation\end{tabular} & \begin{tabular}[c]{@{}l@{}}Short   Description\end{tabular}                                                                                     \\ \hline
Feature                                                                  & $F$                                                             & \begin{tabular}[c]{@{}l@{}}One field  within the input.\end{tabular}                                                                           \\ \hline
Record                                                                   & $R$                                                             & \begin{tabular}[c]{@{}l@{}}A collection of one item of information (e.g. picture, row in datasheet).\end{tabular}                             \\ \hline
Target                                                                   & $T$                                                             & \begin{tabular}[c]{@{}l@{}}The labelled category to be learned. Can be categorical or numeric.\end{tabular}                                   \\ \hline
\begin{tabular}[c]{@{}l@{}}Algorithm\end{tabular}   & $L$                                                             & \begin{tabular}[c]{@{}l@{}}The algorithm used to predict the value of $T$ from the collection of data \\(all features and records).\end{tabular}               \\ \hline
\begin{tabular}[c]{@{}l@{}}Interpretation\end{tabular} & $\I$                                                             & \begin{tabular}[c]{@{}l@{}}A function that takes as its input $F,R,T,$ and $L$ \\and returns a representation of $L$'s logic.\end{tabular}                                           \\ \hline
Explanation                                                           & \multicolumn{1}{c|}{$\E$}                                         & \multicolumn{1}{l|}{\begin{tabular}[c]{@{}l@{}} The human-centric objective for the user to understand $L$ using $\I$.\end{tabular}}                              \\ \hline
Explicitness                                                             & \multicolumn{1}{l|}{}                                         & \multicolumn{1}{l|}{\begin{tabular}[c]{@{}l@{}}The extent  to which $\I$ is understandable to the  intended user.\end{tabular}} \\ \hline
Fairness                                                                 &                                                               & \begin{tabular}[c]{@{}l@{}}The lack  of bias in $L$ for a field   of importance (e.g. gender, age, ethnicity).\end{tabular}                      \\ \hline
Faithfulness                                                             & \multicolumn{1}{l|}{}                                         & \multicolumn{1}{l|}{\begin{tabular}[c]{@{}l@{}}The extent   to which the logic within $\I$ is similar to that of $L$.\end{tabular}}                 \\ \hline
Justification                                                             & \multicolumn{1}{l|}{}                                         & \multicolumn{1}{l|}{\begin{tabular}[c]{@{}l@{}}Why the user should accept $L$'s decision. \\Not necessarily faithful as no connection assumed between $L$ and $\I$.\end{tabular}}                 \\ \hline
Transparency                                                             & \multicolumn{1}{l|}{}                                         & \multicolumn{1}{l|}{\begin{tabular}[c]{@{}l@{}}The connection between $\I$ and $L$ is both explicit and faithful.\end{tabular}}  \\ \hline
\end{tabular}
\caption{Notation and short definition of key concepts of explainability, interpretability, transparency, fairness, and explicitness in this paper.  Concepts of features, records, targets and machine learning algorithms and explanations are also included as they define the key concepts.}
\label{table2}
\end{table}

We define explainability as the ability for the human user to understand the agent's logic. This definition is consistent with several papers that considered the difference between explainability and interpretability within Human-Agent Systems. For example, Doran et al. define explainable systems as those that explains the decision-making process of a model using reasoning about the most human-understandable features of the input data \cite{DoranSB17}. Following their logic, interpretability and transparency can help form explanations, but are only part of the process. Guidotti et al. state that ``an interpretable model is required to provide an explanation" \cite{Guidotti2018}, thus an explanation is obtained by the means of an interpretable model. 
Similarly, Montanavon et al., define explanations as ``a collection of features of the interpretable domain, that have contributed for a given example to produce a decision" \cite{post}. 

Thus, the objective of any system is explainability, meaning it has an explanation $\E$, which is the human-centric aim to understand $L$. An explanation is derived based on the human user's understanding about the connection between $T$ and $R \times F$. The user will create $\E$ based on her understanding of an interpretation  function, $\I$ that takes as its inputs $L$, $R \times F$ and $T$ and returns a representation of the logic within $L$ that can be understood. Consequently, in this paper we refer to explainability of systems as the understanding the human user has achieved from the explanation and do not use this term interchangeably with ``interpretability" and ``transparency". We reserve use of terms ``interpretability" and ``transparency" as descriptions of the agent's logic.  Specifically, we define $\E$ as:
\begin{equation} 
\E = \I(L(R \times F,T))
\label{equ:explain}
\end{equation}

We claim that the connection between $\I$ and $L$, $R$, $F$ and $T$ will also determine the type of explanation that is generated.  A globally explainable model provides an explanation for all outcomes within $T$ taking into consideration $R \times F$, thus using all information in: ${L,R,F,T}$. A locally explainable model provides explanations for a specific outcome, $t \in T$ (and by extension for specific records $r \in R$), using ${L,r,F,t}$ as input. 

We use three additional terms: explicitness, faithfulness and justification to quantify the relationship of $\I$ to $\E$ and $L$ respectively. Following recent work \cite{abs-1806-07538}, we refer to \textit{explicitness} as the level to which the output of $\I$ is immediate and understandable.  As we further explore in the next section, the level of explicitness depends on \emph{who} the target of the explanation is and what is the level of her expertise at understanding $\I$. It is likely that two users will obtain different values for $\E$ even given the same value for $\I$, making quantifying $\I$'s explicitness difficult due to this level of subjectivity. We define \textit{faithfulness}, also previously defined as \textit{fidelity} \cite{4938655,Ribeiro2016}, as the degree to which the logic within $\I$ is similar to that of $L$. Especially within less faithful models, a concept of \textit{completeness} was recently suggested to refer to the ability of $\I$ to provide an accurate description for all possible actions of $L$ \cite{abs-1806-00069}. Given the similarity of these terms, we only use the term faithful due to its general connotation. Justification was previously defined as an explanation about why a decision is correct without any information about the logic about how it was made \cite{biran2017explanation}. According to this definition, justifications can be generated even within non-interpretable systems. Consequently, justification requires no connection between $\I$ and $L$ and no faithfulness. Instead, justification methods are likely to provide implicit or explicit arguments about the correctness of the agent's decision, such as through persuasive argumentation \cite{yetim2008framework}.

In order for a model to be transparent, two elements are needed: the decision-making model must be readily understood by the user, and that explanation must map directly to how the decision is made. More precisely, a transparent explanation is one where the connection between $\E$, $\I$ and $L$ is explicit and faithful as the logic within $\I$ is readily understandable and identical to $L$, e.g. $\I \simeq L$. When a tool or model is used to provide information about the decision-making process secondary to $L$, the system contains elements of interpretability, but not transparency.  

Section \ref{What} discusses the different types of interpretations that can be generated, including transparent ones. Non-transparent interpretations will lack faithfulness, explicitness, or both. Examples include tools to create model and outcome interpretations, feature analysis, visualization methods and prototype analysis. Each of these methods will focus on different parameters within the input, ${R,F,T}$ and their relationship to $L$. Model and outcome interpretation tools create $\I$ without a direct connection to the logic in $L$. Feature Analysis is a method of providing interpretations via analyzing a subset of features $f \in F$. Prototype selection is a method of providing interpretations via analyzing a subset of records $r \in R$. Visualization tools are used to understand the connection between $L$ and $T$ and thus $\I$ takes this interpretable form. 

\begin{figure}
\centering
\includegraphics[width=4.5in]{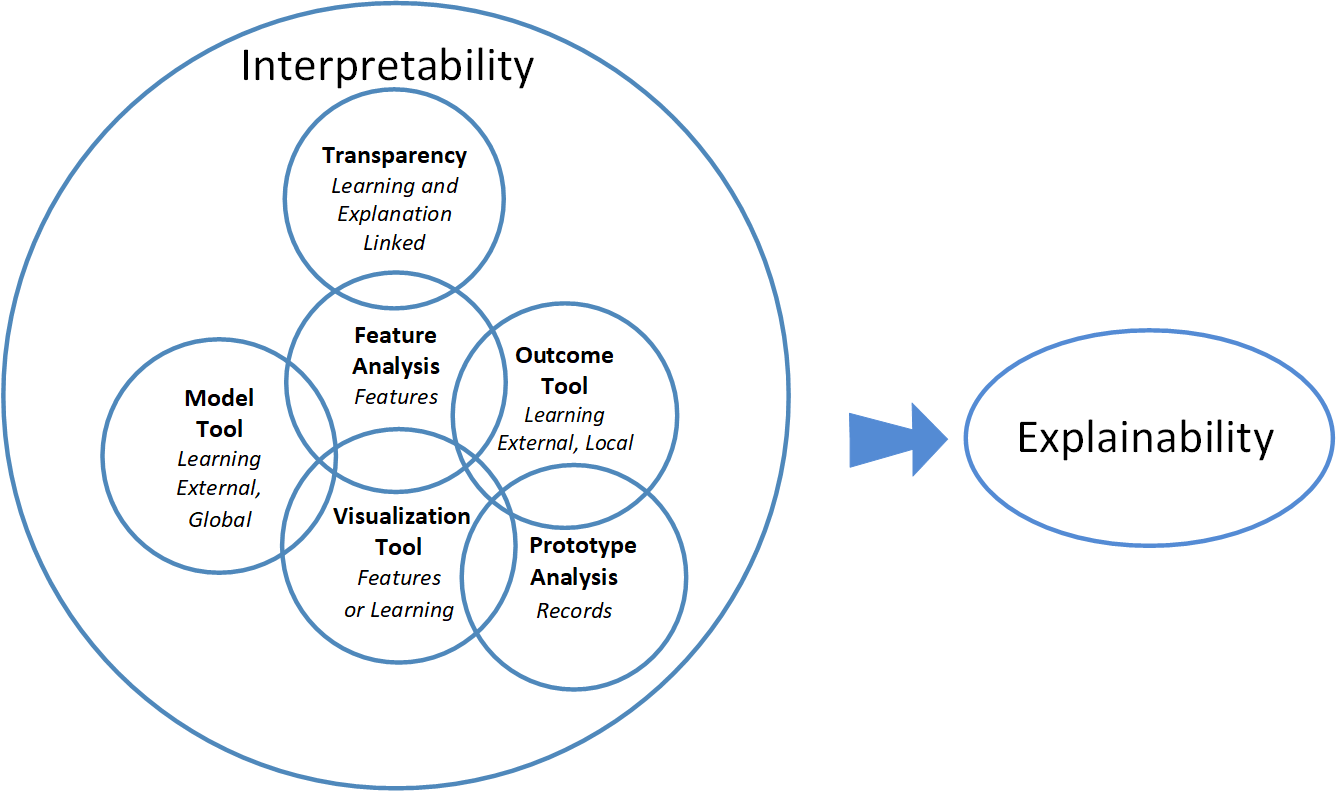}
\caption{A Venn Diagram of the relationship between Explainability, Interpretability and Transparency. Notice the centrality of Feature Analysis to 4 of the 5 interpretable elements.}
\label{Figure1} \label{fig::example0}
\end{figure}

To help visualize the relationship between explainability, interpretability and transparency, please note Figure \ref{Figure1}. Note that interpretability includes six methods, including transparent models, and also the non-transparent possibilities of model and outcome tools, feature analysis, visualization methods, and prototype analysis.  In the figure, interpretability points to the objective of explainability to signify that interpretability is a means for providing explainability, as per these terms' definitions in Table \ref{table2}. Note the overlaps within the figure. Feature analysis can serve as a basis for creating transparent models, on its own as a method of interpretability, or as a interpretable component within model, outcome and visualization tools. Similarly, visualization tools can help explain the entire model as a global solution or as a localized interpretable element for specific outcomes of $t \in T$. Prototype analysis uses $R$ as the basis for interpretability, and not $F$, and can be used for visualization and/or outcome analysis of $r \in R$. We explore these points further in Section \ref{What}. 

The level of interpretability and transparency needed within an explanation will be connected to either hard or soft-constraints defined by the user's requirements. At times, there may be a hard-constraint based on a legal requirement for transparency, or a soft-constraint that transparency exist in cases where one suspects that the agent made a mistake or does not understand why the agent chose one possibility over others \cite{gregor1999explanations,Guidotti2018,schank1986explanation,van198511,vlek2016method}. Explainability can be important for other reasons, including building trust between the user and system even when mistakes were not made \cite{chen2014situation}-- something we now explore.

\section{\emph{Why} a Human-Agent System should be Explainable? }\label{Why}

We believe that the single most important question one must ask about this topic is \emph{Why} we need an explanation, and how important it is for the user to understand the agent's logic. In answering this question one must establish whether a system truly needs to be explainable. We posit that one can generalize the need for explainability with a taxonomy of three levels: 
\begin{enumerate}
\item Not helpful
\item Beneficial
\item Critical
\end{enumerate}

Adjustable autonomy is a well-established concept within human-agent and human-robot groups that refers to the amount of control an agent/robot has compared to the human-user \cite{goodrich2001experiments,scerri2001adjustable,yanco2004classifying}.  Under this approach, the need for explainability can be viewed as a function of the degree of cooperation between the agent to the human user. Assuming the agent is fully controlled by the human operator (e.g. teleoperated), then no explainability is needed as the agent is fully an extension of the human participant. Conversely, if the robot is given full control, particularly if the reason for the decision is obvious (a recommendation agent gives advice based on a well-established collaborative filtering algorithm), it again serves to reason that no explainability is needed. Additionally, Doshi-Velez and Kim pointed out that an explanation at times is not needed if there are no significant consequences for unacceptable results or the  agent's decision is universally accepted and trusted \cite{doshi2017towards}.

At the other extreme, many Human-Agent Systems are built whereby the agent's role is to support a human's task. In many of these cases, we argue that the agent's explanation is a critical element within the system. The need for an agent to be transparent or to explicitly and faithfully explain its actions is tied directly to task execution. For example, Intelligent Tutoring Systems (ITS) typically use step-based granularities of interaction whereby the agent confirms one skill has been learned or uses hints to guide the human participant \cite{vanlehn2011relative}. The system must provide concrete explanations for its guidance (called \textit{hints} in ITS terminology) to better guide the user. Similarly, explanations form a critical component of many negotiation, training, and argumentation systems \cite{rahwan2003towards,rosenfeld2016providing,rosenfeld2016negochat,sierhuis2003human,traum2003negotiation}. For example, effective explanations might be critical to aid a person in making the final life-or-death decision within Human-Agent Systems \cite{sierhuis2003human}. Rosenfeld's et al.'s NegoChat-A negotiation agent uses arguments to present the logic behind its position \cite{rosenfeld2016negochat}. Traum et al. explained the justification within choices of their training agent to better convince the trainee, as well as to teach the factors to
look at in making decisions \cite{traum2003negotiation}. Rosenfeld and Kraus created agents that use argumentation to better persuade people to engage in positive behaviors, such as choosing healthier foods to eat \cite{rosenfeld2016providing}. Azaria et al. demonstrate how an agent that learns the best presentation method for proposals given to a user  improves their acceptance rate \cite{azaria2014agent}. Many of these systems can be generally described as Decision Support Systems (DSS). A DSS is typically defined as helping people make semi-structured decisions  requiring some human judgment and at the same time with some agreement on the solution method \cite{Adam2008}.  An agent's effective explanation is critical within a DSS as the system's goal is providing the information to help facilitate improved user decisions.

A middle category in our taxonomy exists when an explanation is beneficial, but not critical. The Merrian-Webster dictionary defines beneficial as something that ``produces good or helpful results"\footnote{https://www.merriam-webster.com/dictionary/benefit}.  In general, the defining characteristic of explanations within this category is that they are not needed in order for the system to behave optimally or with peak efficiency. 

To date, many reasons have been suggested for making systems explainable \cite{CHI2018,DoranSB17,doshi2017towards,gregor1999explanations,Guidotti2018,Lipton16a,sormo2005explanation}:

\begin{enumerate}
\item To justify its decisions so the human participant can decide to accept them (provide control)
\item To explain the agent's choices to guarantee safety concerns are met
\item To build trust in the agent's choices, especially if a mistake is suspected or the human operator does not have experience with the system
\item To explain the agent's choices to ensure fair, ethical, and/or legal decisions are made
\item Knowledge / scientific discovery 
\item To explain the agent's choices to better evaluate or debug the system in previously unconsidered situations
\end{enumerate}

The importance of these types of explanations will likely vary greatly across systems. If the user will not accept the system without this explanation, then a critical need for explainability exists. This can particularly be the case in Human-Agent Systems where the agent supports a life-or-death task, such as search and rescue or medical diagnostic systems, where ultimately the person is tasked with the final decisions \cite{jennings2014human}. In these types of tasks the explanation is critical to facilitate a person's decision whether to accept the agent's suggestion and/or to allow that person to decide if safety concerns are met, such as a patient's health or that of a person at-risk in a rescue situation. In other situations, explanations are beneficial for the overall function of the human-agent system, but are not critical. 

One key and common example where explanations can range in significance from critical to beneficial are situations where explanations help instill trust. Previous work on trust, within people in a work situation, identified two types of trust that develop over time, ``knowledge-based" and ``identification-based" \cite{inbook}.  Of the two types of trust, they claim that knowledge-based trust requires less time, interactions and information to develop as it is grounded primarily in the other party's predictability. Identification-based trust requires a mutual understanding about the other's desires and intention and requires more information, interactions and time to develop.  

We posit that previous work has focused on elements of this trust model in identifying what types of explanations are necessary to foster this type of trust within Human-Agent Systems. Following our previous definitions of interpretability and transparency, it seems that the former type of interpretable elements may be sufficient for knowledge-based definitions of trust, while transparent elements are required for identification-based models. When a person has not yet developed enough positive experience with the agent she interacts with, both knowledge-based and identification based trust are missing. As it has been previously noted that people are slow to adopt systems that they do not understand and trust and ``if the users do not trust a model or a prediction they will not use it." \cite{Ribeiro2016}, even providing a non-transparent interpretable explanation will likely help instill confidence about the system's predictability, thus facilitating the user's knowledge-based trust in the system. Ribeiro et al. demonstrate how interpretability of this type is important for identifying  models that have high accuracy for the wrong reasons \cite{Ribeiro2016}. For example, they show that  text classification often are wrongly based on the heading rather than the content. In contrast, image classifiers that capture  the main part of the image in a similar manner to the human eye, install a feeling that the model is functioning correctly even if accuracy is not particularly high.  

However, it has been claimed that  when the person suspects the agent has made a mistake and/or is unreliable then the agent should act with transparency, and not merely be interpretable, as explanations generated from transparent methods will aid the user to trust the agent in the future \cite{XAIP2017}. In extreme cases, if the user completely disregards the agent, then the human-agent system breaks down, making transparent explanations critical to help restore trust. Furthermore, explanations based on $L$'s transparency  may be needed to help facilitate the higher level of identification-based trust. Only transparent interpretations directly link $L$ and $\I$ thus providing full information about the agent's intention.  We suggest that designers of systems that require this higher level of trust, such as health-care \cite{crockett2016data}, recommender systems \cite{knijnenburg2012explaining}, planning \cite{XAIP2017} and human-robot rescue systems \cite{rosenfeld2017intelligent,salem2015would} should be transparent, and not merely interpretable. 

Other types of explanations are geared towards people beyond the immediate users of the system.  Examples of these types of explanations include those designed for legal and policy experts to confirm that the decisions / actions of agent fulfill legal requirements such as being fair and ethical \cite{DoranSB17,doshi2017towards,dwork2012fairness,Garfinkel2017,Guidotti2018}.  Both the EU and UK governments have adopted guidelines requiring agent designers to provide users information about agents' decisions. In the words of the EU's ``General Data Protection Regulation" (GDPR), users are legally entitled to obtain ``meaningful explanation of the logic involved" of these decisions. Additional legislation exists to ensure that agents are not biased against any ethnic or gender groups \cite{doshi2017towards,Guidotti2018} such that they demonstrate fairness \cite{dwork2012fairness}. Similarly, the ACM has published guidelines for algorithmic accountability and transparency \cite{Garfinkel2017}. 
The system's explanation is not here critical for effective performance of the agent, but instead to confirm that a secondary legal requirement is being met.  

Explanations geared beyond the immediate user can also be those geared for researchers to help facilitate scientific knowledge discovery \cite{doshi2017towards,Guidotti2018} or for system designers to evaluate or test a system \cite{doshi2017towards,sormo2004explanation,sormo2005explanation}. For example, a medical diagnostic system may work with peak efficiency exclusively as a black box, and users may be willing to rely on this black box as the agent is trusted due to an exemplary historical record. Nonetheless, explanations can be still be helpful for knowledge discovery to help researchers understand gain understanding of various medical phenomena.  Explainability has also been suggested as being necessary for properly evaluating a system or for the agent's designer to confirm that  the system is properly functioning, even within situations that were not considered when the agent was built.   For example, Doshi-Velez and Kim claimed that due to the inherent inability to quantify all possible situations, it is impossible for a system designer to evaluate an agent in all possible situations \cite{doshi2017towards}. Explanations can be useful in these situations to help make evident any possible gaps between an agent's formulation and implementation and its performance. In all cases, the explanation is not geared to the end-user of the system, but rather to an expert user who requires the explanation for a reason beyond the day-to-day operation of the system. 

As we have shown in this section, the question about explainability can be divided into questions about its necessity, e.g.  not necessary, beneficial or critical, which is directly connected to the objective of that explanation.  From a user-perspective, the primary objective of the explanation is related to factors that help her use the system, and particularly elements that help foster trust.  In these cases, a system may need to be transparent, even if this level of explanation entails a sacrifice of the system's performance. We further explore this possibility and relationship in Section \ref{What}. 
At times, explanations are needed or beneficial for entities beyond the typical end-user such as for the designer, researcher or legal expert.  As the objective of explanations of this type is different, it stands to reason that the type of explanation may be fundamentally different based on \emph{whom} the target is for this information, something we address in the next section.  This in turn may impact the type of interpretation the agent must present, something we explore in Section \ref{What}. 

\section{\emph{Who} is the Target of the Explanation?}
\label{Who}
The type of interpretable element needed to form the basis of the explanation is highly dependent on the question of \emph{who} the explanation is for. We suggest three possibilities:
\begin{enumerate}
\item Regular user
\item Expert user
\item External entity
\end{enumerate}

The level of explanation detail needed depends on \emph{why} that Human-Agent Systems needs the user to understand the agent's logic (Section \ref{Why})  and how the explanation has been generated (Section \ref{What}). If the need for explanation is for legal purposes, then it follows that legal experts need the explanation, and not the regular user. Similarly, it stands to reason that the type of explanation that is given should be directed specifically to this population. If the purpose of the explanation is to support experts' knowledge discovery, then it stands to reason that the explanation should be directed towards researchers with knowledge of a specific problem. In these cases, the system might not even need to present their explanations to the regular users and may thus only focus on presenting information to these experts.  Most systems will still likely benefit by directing explanations to the regular users to help them better understand the system's decisions, thus aiding in their acceptance and/or trust. In these cases, the system should be focused on providing justifications in addition to providing the logic behind their decisions through arguments \cite{rosenfeld2016providing,yetim2008framework} and/or through Case Based Reasoning \cite{corchado2003constructing,kim2014bayesian,kwon2004applying} that help reassure the user about the correctness of the agent's decision.

The same explanation might be considered as extremely helpful by a system developer, but considered useless by a regular user. Thus, the expertise level of the target will play a large part of defining an explanation and how explicit $\I$ is. Deciding on how to generate and present $\I$ will be covered in later sections.

Similarly, what level of detail constitutes an adequate explanation likely depends on precisely how long the user will study the explanation. If the goal is knowledge discovery and/or complying with legal requirements, then an expert will likely need to spend large amounts of time meticulously studying the inner-workings of the decision-making process. In these cases, it seems likely that great amounts of detail regarding the justification of the explanations will be necessary.  If a regular user is assumed, and the goal is to build user trust and understanding, then shorter, very directed explanations are likely more beneficial.  This issue touches upon a larger issue about the danger additional information may overload a given user \cite{shrot2014crisp}.

At times, the recipient of the explanation is not the user directly interacting with the system.  This is true in cases where explanations are mandated by an external regulative entity, such as is proposed by the EU's GDPR, regulation. In this case, the system must follow explanation guidelines provided by the external entity. In contrast, developers providing explanations to users will typically follow different guidelines, such as user usability studies. As these two types of explanations are not exclusive, it is possible that the agent will generate multiple types of explanations for the different targets (e.g. the user and the regulator entity). In certain types of systems, such as security systems, multiple potential targets of the explanation  also exist. Vigano and Magazzeni explain that security systems have many possible targets, such as the designer, the attacker, and the analyst \cite{vigano2018explainable}. Obviously an explanation provided for the designer can be very dangerous in the hands of an attacker. Thus, aside from the question of how ``helpful" an explanation is for a certain types of user, one must consider what the implications of providing an unsuitable explanation are.  In these cases, the explanation must be provided for a given user while also considering the implications on the system's security goals. 

\section{\emph{What} Interpretation can be Generated?}\label{What}
Once we have established the \emph{why} and  \emph{who} about explanations, a key related question one must address is \emph{what} interpretation can be generated as the basis for the required explanation. Different users will need different types of explanations, and the interpretations required for effective explanations will differ accordingly \cite{vigano2018explainable}. We posit that six basic approaches exist as to how interpretations can be generated:
\begin{enumerate}
\item Directly from a transparent machine learning algorithm 
\item Feature selection and/or analysis of the inputs
\item Using an algorithm to create a post-hoc model tool
\item Using an algorithm to create a post-hoc outcome tool
\item Using an interpretation algorithm to create a post-hoc visualization of the agent's logic
\item Using an interpretation algorithm to provide post-hoc support for the agent's logic via prototypes
\end{enumerate}

\begin{figure}
\label{Figure2}
\centering
\includegraphics[width=3.3in]{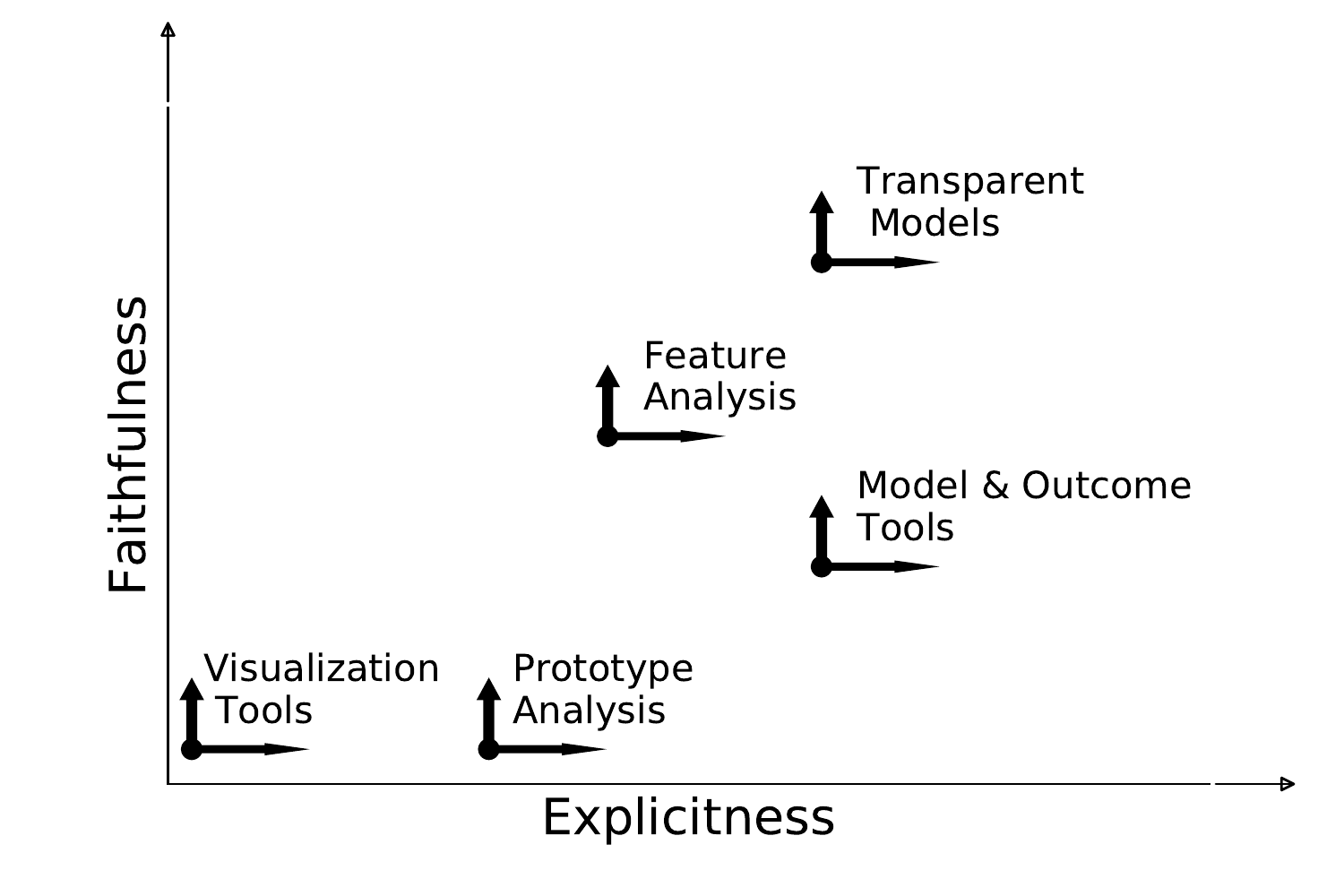}
\caption{Faithfulness versus explicitness within the six basic approaches for generating interpretations}
\label{fig::Explicit-Faithful}
\end{figure}

In Figure \ref{fig::Explicit-Faithful} we describe how these  various methods for generating interpretations have different degrees of faithfulness and explicitness.  Each of these methods contains some level of trade-off between their explicitness and faithfulness.  For example, as described in Section \ref{definitions}, transparent models are inherently more explicit and faithful than other possibilities. Nonetheless, we present this figure only as a guideline, as many implementations and possibilities exists within each of these six basic approaches. These differences will impact the levels of both faithfulness and explicitness, something we indicate via the arrows pointing to both higher levels of faithfulness and explicitness for a specific implementation.  

\subsection{Generating Transparent Interpretations Directly from Machine Learning Algorithms}
\label{What-transparent}
\begin{figure}
\label{Figure3}
\centering
\includegraphics[width=3.3in]{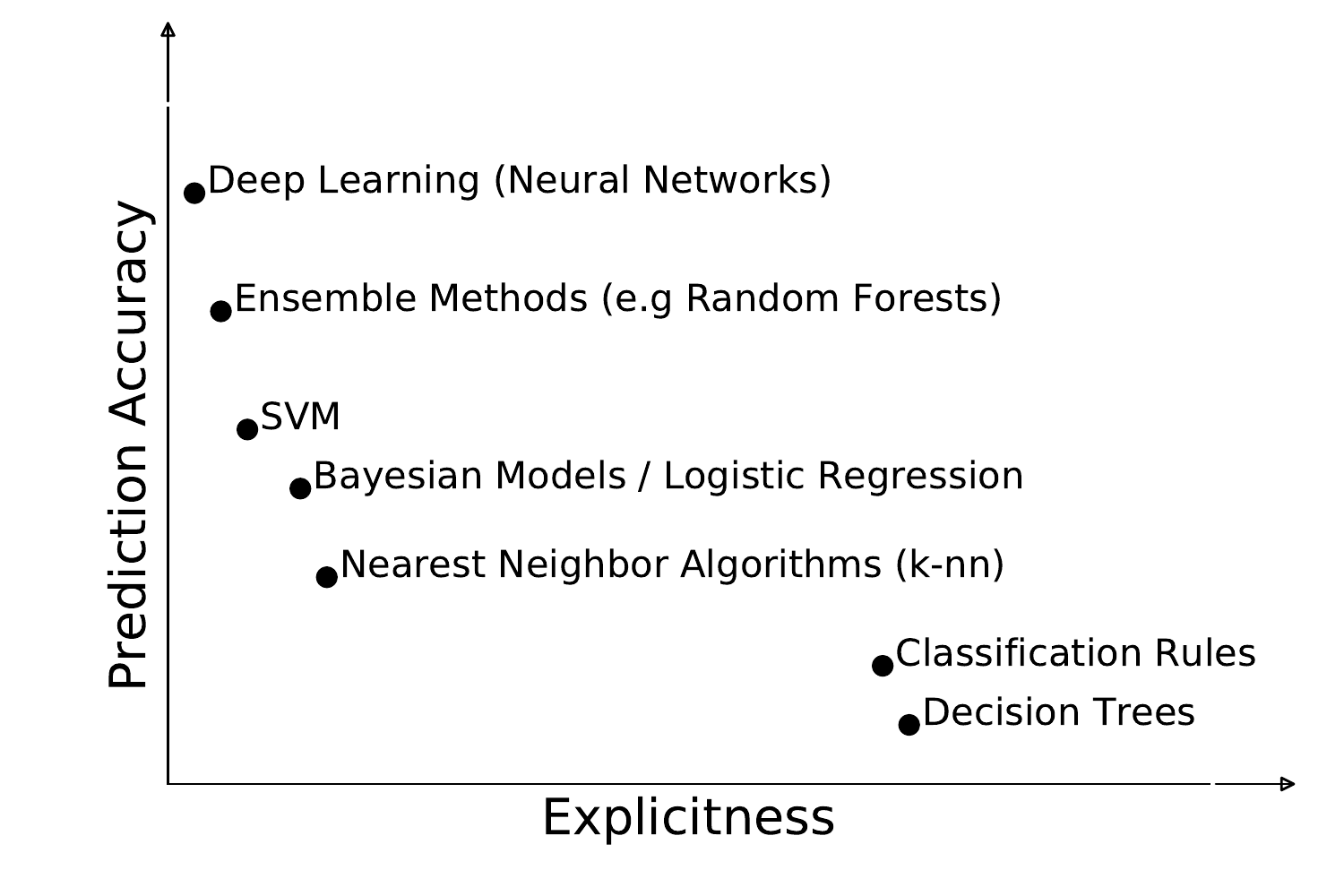}
\caption{Typical trade-off between prediction accuracy versus explicitness}
\label{fig::Explicit-Predict}
\end{figure}

The first approach, and the most explicit and faithful method, is to generate $\I$ directly from the output of the machine learning algorithm, $L$. These types of interpretations can be considered ante-hoc, or ``before this" (e.g. an explanation is needed), as the this type of connection between $\I$ and $L$ facilitates providing interpretations at any point, including as the task is being performed \cite{ante-2018,holzinger2017we}.
These transparent algorithms, often called white box algorithms,  include decision trees, rule-based methods, k-nn (k-nearest neighbor), Bayesian and logistic regression \cite{dreiseitl2002logistic}. As per our definitions in Section \ref{sec:definitions}, these algorithms have not been designed for generating interpretations, but can be readily derived from the understandable logic inherent in the algorithms. As we explain in this section,  all of these algorithms are faithful, and are explicit to varying degrees. A clear downside to these approaches is that one is then limited to these machine learning algorithms, and/or a specific algorithmic implementation. It has been previously noted that an inverse relationship often exists between machine learning algorithms' accuracy and their explainability \cite{Guidotti2018,gunning2017explainable}. Black box algorithms, especially deep neural networks but including other less explainable algorithms such as ensemble methods and support vector machines, are often used due to their exceptional accuracy on some problems. However, these types of algorithms are difficult to glean explicit interpretations from and are typically not transparent \cite{dreiseitl2002logistic}. Figure \ref{fig::Explicit-Predict} is based on previous work \cite{Explain2018,gunning2017explainable} and quantifies the general relationship between algorithms' explicitness and accuracy. This figure describes the relationships as they stand at the time the paper is written, and may change as algorithmic solutions develop and evolve. Additionally, this figure may be somewhat over-simplified, as we now describe.

Decision trees are often cited to be the most understandable (e.g. explicit) \cite{Explain2018,doshi2017towards,Freitas2014,gunning2017explainable,quinlan1986induction}. The hierarchical structure inherent in decision trees yields itself to understanding which attributes are most important, of second-most importance, etc. \cite{Freitas2014}. Furthermore, assuming the size of the tree is relatively small due to Occam's Razor \cite{murphy1993exploring}, the if-then rules that can be derived directly from decision trees are both particularly explicit and faithful \cite{Freitas2014,RosenfeldSGBHL14}.  

However, in practice not all decision trees are easily understood. Large decision trees with hundreds of nodes and leaves are often more accurate than smaller ones, despite the assumption inherent within Occam's Razor \cite{murphy1993exploring}. Such trees are less explicit, especially if they contain many attributes and/or multiple instances of nodes using the same attribute for different conditions. Assuming the decision tree is too large to fully understand  (e.g. thousands of rules) \cite{hara2016making} and/or overfitted due to noise in the training data \cite{Freitas2014}, it will lose its explicitness. One approach to address this issue is suggested by Last and Maimon \cite{last2004compact} where they reason about the added value of added attributes versus the complexity they add, facilitating more explicit models. 

Classification rules \cite{clark1989cn2,michalski1999learning} have also been suggested as a highly explicit machine learning model \cite{Explain2018,Freitas2014,gunning2017explainable}. As is the case with decision trees, the if-then rules within such models provide faithful interpretations and are potentially explicit. The flat, non-hierarchical structure in such models can be an advantage in allowing the user to focus on individual rules separately which at times has been shown to be advantageous \cite{clancey1983epistemology,Freitas2014}. However, in contrast to decision trees, this structure does not inherently give a person insight as to the relative importance of the rules within the system. Furthermore, conflicts between rules need to be handled, often through an ordered rule-list, adding to the model's complexity and reducing its level of explicitness. 

Nearest neighbor algorithms, such as k-nn, can potentially be transparent machine learning models as they can provide interpretations based on the similarity between an item needing interpretation and other similar items. This is reminiscent of the picture classification example in Section \ref{sec:definitions} as the person is actually performing an analysis similar to k-nn in understanding why certain pictures are similar. This process is also similar to the logic within certain Case Based Reasoning algorithms which often also use logic akin to k-nn algorithms to provide an interpretation for why two items are similar \cite{sormo2005explanation}.  However, as has been previously pointed out, these interpretations are only typically explicit if k is kept small, e.g. k=1 or close to 1 \cite{sormo2005explanation}. Furthermore, k-nn is a lazy model that classifies each new instance separately. As such, every instance could potentially have a different ``interpretation", making this a local interpretation. In contrast, both decision trees and rule-based systems construct general rules that are to be applied across all instances \cite{Freitas2014}. In addition, if the number of attributes in the dataset are very large, it might be difficult for a person to appreciate the similarities and differences between different instances again reducing the explicitness of the model. 

Bayesian network classifiers have also been suggested as another transparent machine learning model. Knowing the probability of a successful outcome is often needed in many applications, something that probabilistic models, including Bayesian models, excel at \cite{bellazzi2008predictive}. Bayesian models have been previously suggested to be the most transparent of these types of models as each attribute can be independently analyzed and the relative strength of that attribute be understood \cite{Freitas2014}. This approach is favored in many medical applications for this reason \cite{zupan2000machine,kononenko1993inductive,lavravc1999selected}. More complex, non-na\"ive Bayesian models can be constructed \cite{cheng1999comparing} although one may then potentially lose both model accuracy and transparency. 

Similar to Bayesian models, logistic regression also outputs outcome probabilities by fitting the output of its regression model to values between 0 and 1. The logit function inherent in this model is also constructed from probabilities-- here in the form of log-odds / odds-ratios. This makes this model popular for creating medical applications \cite{bagley2001logistic,katafigiotis2018stone}. At times, the interpretations that can be generated by these relationships are explicit \cite{dreiseitl2002logistic}.  

Support Vector Machines (SVM) are based on finding a hyperplane to separate between different instances and are potentially explicit, particularly if a linear kernel is used \cite{bellazzi2008predictive}. Once again, if many attributes exist in the model, the explicitness of the model might be limited even if a linear kernel is used. An SVM becomes even less explicit if more complex kernels are used including RBF and polynomial kernels. As is the case with the last three of these algorithms (SVM, k-nn and Bayesian), feature selection / reduction could significantly help the explicitness of the model, something we explore in the next section. 

As no one algorithm provides both high accuracy and explicitness, it is important to consider new machine learning algorithms that include explainability as a consideration within the learning algorithm. One example of this approach is work by Kim, Rudin and Shah, who have suggested a Bayesian Case Model for case-based reasoning \cite{kim2014bayesian}.  Another example is introduced by Lou et al. \cite{LouCG12}. Their generalized additive models (GAMs) combine univariate models called shape functions through a linear function. On one hand, the shape functions can be arbitrarily complex, making GAMs more accurate than simple linear models. On the other hand, GAMs do not contain any interactions between features, making them more explicit than black box models. Lou et al. also suggested adding selected terms of interacting pairs of features to standard GAMs \cite{lou2013accurate}. This method increases the accuracy of the models, while maintaining better explicitness than black box methods. Caruana et al. propose a extension of the GAM, GA$^2$M, which considers pairwise interactions between features and provide a case study showing its success in accurately and transparently explaining a health-care dataset \cite{Caruana2015}.

We believe these approaches are worthy of further consideration and provide an important future research area as new combinations of machine learning algorithms that provide both high accuracy and explainability could potentially be developed. Several of these methods use an element of feature analysis as the basis of their transparency \cite{Caruana2015,last2004compact,LouCG12,lou2013accurate}. In general, feature selection can be a critical element in creating transparent and non-transparent interpretations, as we now detail.

\subsection{Generating Interpretations from Feature Selection / Analysis}
\label{What-feature analysis}
A second approach to create the interpretation, $\I$, is through performing feature selection and/or feature analysis of all features, $F_1 \ldots F_i$, before or after a model has been built. Theoretically, this approach can be used alone and exclusively to generate interpretations within the non-transparent ``black box" algorithms, or in conjunction with the above ``white box" algorithms to help further increase their explicitness. Feature selection has long been established as an effective way of building potentially better models which are simpler and thus better overcome the curse of dimensionality \cite{guyon2003introduction}. Additionally, models with fewer attributes are potentially more explicit as the true causal relationship between the dependent and independent variables is clearer and thus easier to present to the user \cite{Kononenko99}. The strong advantage of this approach is that the information presented to the user is generated directly from the mathematical relationship between a small set of features and the target being learned.  

Three basic types of feature selection approaches exist: filters, wrappers, and embedded methods.  We believe that filter methods are typically best suited for generating explicit interpretations as the analysis is derived directly from the data without any connection to a specific machine learning model \cite{guyon2003introduction,Saeys2007survey}. Univariate scores such as information gain or $X^2$ can be used to evaluate each of the attributes independently. Either the top $n$ features could then be selected or only those with a score above a previously defined threshold. The user's attention could then be focused on relationships between these attribute, facilitating explicitness. Multivariate filters, such as CFS \cite{hall1999correlation} allow us to potentially discover interconnections between attributes. The user's attention could again then be focused on this small subset of features with the assumption that interrelationships between features have become more explicit. Previous work by Vellido et al. \cite{vellido2012making} recommends using principals component analysis (PCA) to generate interpretations. Not only does PCA reduce the number of attributes needing to be considered, but the new features generated by PCA are linear combinations of the original ones. As such, the user could understand an explanation based on these interrelationships, especially if the both the number and size of these derived features are small. 

As filter methods are independent of the machine learning algorithm used, it has been suggested that this approach can be used in conjunction with black box algorithms to make them more explicit \cite{vellido2012robust}.  One example is previous work that used feature selection to reduce the number of features from nearly 200 to 3 before  using a neural network for classification \cite{vellido2012robust}. As neural networks are becoming increasing popular due to their superior accuracy in many datasets, we believe this is a general approach that is worth consideration to help make neural networks more explicit.

\subsection {Tools to Generate Model Interpretations Independently from L}
\label{What-Model outcome}
The above methods are faithful in that the transparent algorithms and feature analysis is done in conjunction with $L$.  However, other approaches exist that create $\I$ as a process independent of the logic within $L$. In the best case, $\I$ does faithfully approximate the actual and complete logic within $L$, albeit found differently, and thus represents a form of reverse-engineering version of the logic within $L$ \cite{Augasta2012}. Even when $\I$ is not 100\% faithful, the goal is to be as faithful and explicit as possible, making these approaches a type of metacognition process, or reasoning about the reasoning  process (e.g. $L$)  \cite{cox2011metareasoning}. A key difference within the remaining approaches in this section is that $\I$ is created through an analysis after the $L$'s learning has been done, something referred to as postprocessing \cite{Strumbelj2010} or post-hoc analysis \cite{Lipton16a,post}. Examples of post-hoc approaches that we consider in the remainder of this section include: model and outcome interpretations, visualization, and prototyping similar records. 

While disconnecting the $L$ and $\I$ can lead to a loss of faithfulness, it can lead to other benefits and challenges. Designing tools that focus on $\I$ could potentially lead to very explicit models, something we represent in Figure \ref{fig::Explicit-Faithful}. Additionally, interpretations that are derived directly from the machine learning algorithm or the features are strongly restricted by the nature of the algorithm / features. In contrast, interpretations that are created in addition to the decision-making algorithm can be made to comply with various standards. For example, Miller demonstrates how  interpretations are often created by the same people that develop the system. They tend to generate explanations that are understandable to software designers, but are not explicit for the system's users \cite{miller2017explanation}. He suggests using insights from the social sciences when discussing explainability in AI. Other factors, such as legal and practical considerations might limit researchers as to what constitutes a sufficient explanation. For example, as these tools disconnect the logic in $\I$ from $L$, they cannot guarantee the fairness of the agent's decision which may be a critical need and even require transparency (see Section \ref{Why}). 

The first possibility creates a ``model interpretation tool" that is used to explain the logic behind $L$'s predictions for all values of $T$ given all records, $R$. A group of these approaches create simpler, transparent decision trees or rules secondary to $L$. While these approaches will have the highest level of explicitness, they will generally lack faithfulness.  For example, Frosst \cite{frosst2017distilling} presents a specific interpretation model for neural networks in an attempt to resolve the tension between the generalization of neural networks and the explicitness of decision trees. They show how to use a deep neural network to train a decision tree. The new model does not perform as well as a neural network, but is explicit. Many other approaches have used decision trees to provide explanations for neural networks \cite{Boz2002,Craven1995,KRISHNAN1999}, decision rules \cite{arbatli1997rule,Augasta2012,craven1994using,Kahramanli,zhou2003extracting} and a combination of genetic algorithms with decision trees or rules \cite{arbatli1997rule,4938655,Mohamed2011}. Similarly, decision trees \cite{Chipman_makingsense,Domingos1998,zhou2016interpreting} and decision rules \cite{deng2014interpreting,hara2016making,4167900,tan2016tree} have been suggested to explain tree ensembles. 

Some explanations secondary to $L$ are generated by using feature analysis and thus are most similar to the approaches in the previous section. One example of these algorithms is   SP-LIME, which provides explanations that are independent of the type of machine learning algorithm used \cite{Ribeiro2016}. It is noteworthy that  SP-LIME includes feature engineering as part of its analysis, showing the potential connection between the second and third approaches. The feature engineering in  SP-LIME tweaks examples that are tagged as positive and observes how changing them affects the classification. A similar method has been used to show how Random Forests can be made explainable \cite{tolomei2017interpretable,whitmore2018explicating}. The Random Forest can be considered a black box that determines the class of a given feature set. $L$'s interpretabity is obtained by determining how the different features contribute to the classification of a feature set \cite{tolomei2017interpretable}, or even which features should be changed, and how, in order to obtain a different classification \cite{whitmore2018explicating}. This type of interpretation is extremely valuable. For example, consider a set of medical features, such as weight, blood pressure, age etc. and a model to determine heart attack risk. Assume that for a specific feature set the model classifies the patient as high risk. The model's interpretation facilitates knowing what parameters need to change in order to change the prediction to low risk.

\subsection {Tools to Generate Outcome Interpretations Independently from L}
\label{What-outcome explanation}
The second possibility for creating interpretations independently from $L$ creates an ``outcome explanation" that is localized and explains the prediction for a given instance $r \in R$ and its prediction, $t \in T$.  It has been claimed that feature selection approaches are useful for obtaining a general, global understanding of the model, but not for specific classifications of an instance, $t$. Consequently, they advocate using local interpretations \cite{Baehrens2010}.   One example is an approach that uses vectors which are constructed independently of the learning algorithm for generating localized interpretations \cite{Baehrens2010}. Another example advocates using coalition game theory to evaluate the effect of combinations of features for predicting $t$ \cite{Strumbelj2010}. Work by Lundberg and Lee present a unified framework for interpreting
predictions using Shapley game theoretic functions \cite{lundberg2017unified}. Certain algorithms have both localized and global versions. One example is the local algorithm LIME and its global variant, SP-LIME \cite{Ribeiro2016}.

\subsection {Algorithms to Visualize the Algorithm's Decision}
\label{What-visualization}
While the explanations in the previous sections focused on ways a person could better understand the logic within $L$, visualization techniques typically focus on explaining how a subset of features within $F$ are connected to $L$. However, the level of explicitness within visualization is lower than that of feature selection and model and outcome interpretations. This is because feature selection and model and outcome interpretations all aim to understand the logic within $L$, thus giving them relatively higher level of faithfulness and explicitness. As visualization tools do not focus on understanding the logic within $L$, they are less faithful than feature analysis methods that do, and at times the level of understanding they provide is not high, especially for regular users.

Overall, many of these approaches seem to have the primary goal of justification for a specific outcome of $L$ and are not focused on even localized interpretations of $L$'s logic. As justification is more concerned with persuading a user that a decision is correct than providing information about $L$'s logic \cite{biran2017explanation}, it seems that justification methods likely have the least amount of faithfulness, as there is no need to make any direct connection between $\I$ and $L$.  Consistent with this aim, work by Lei, Barzilay and Jaakkola generated rationales, which they defined as justifications for an agent's local decision through creating a visualization tool that highlighted which sections of text, e.g. $f \in F$, were responsible for making a specific classification \cite{lei2016rationalizing}. 

Consider explanations that can potentially be generated within image classification, a task many visualization tools address \cite{fong2017interpretable,guo2010novel,Simonyan2013DeepIC,xu2015show,zhou2016learning}. A visualization tool will typically identify the portion of the picture (a subset of $F$) that was most responsible for yielding a prediction, $T_k$. However, typical visualizations, such as those generated by saliency masks, class activation mapping, sensitivity analysis and partial dependency plots all only focus on highlighting important portions of input, without explaining the logic within the model, and the output is often hard for regular users to understand. Nonetheless, these approaches are useful in explaining high accuracy, low-explicitness machine learning algorithms, particularly neural networks, often within image classification tasks.

Saliency maps are a visualizations that identify important, e.g. salient, objects which are groups of features \cite{xu2015show}. In general, saliency can be defined as identifying the region of an image $r \in R$,  that $L(r\times F,T_k)$ will identify \cite{fong2017interpretable}. For example, a picture may include several items, such as a person, house and car.  $r$ can represent the car and $L(r\times F,T_k)$ is used to properly identify it ($T_k$). Somewhat similar to the previous types of explanations, these salient features could then generate a textual explanation of an image. For example, Xu et al. focused on identifying objects within a deep neural network (CNN) for picture identification to automatically create text descriptions \cite{xu2015show} for a given picture (outcome description). Kim et al. created textual explanation for neural networks of self-driving cars \cite{kim2018textual}. More generally, saliency masks can be used to identify the  $r*n$ areas that represent the $t*n$ targets that were identified in the picture \cite{fong2017interpretable,guo2010novel,Simonyan2013DeepIC,xu2015show,zhou2016learning}. They generally use the gradient of the output corresponding to the each of the targets with respect to the inputted features \cite{Lipton16a}. While earlier works constrained the neural network to provide this level of explicitness \cite{zhou2016learning}, recent works provide visual explanations without altering the structure of the neural network \cite{fong2017interpretable,hu2018explainable,selvaraju2017grad}. Still, serious concerns exist that many of these visualizations are too complex for regular users and thus reserved for experts, as some of these explanations are only appropriate for people researching the under-workings of the algorithm to diagnose and understand mistakes \cite{selvaraju2017grad}. 

Neural activation is a visualization for the inspection of neural networks that help focus a person  to what neurons are activated with respect to particular input records. As opposed to the previous visualizations that focus on $F$ and $R$, this visualization helps provide an understanding about neural networks' decisions making them less of a black box. Consequently, these approaches provide interpretation and not justification and are more faithful. For example, work by Yosinski et al. \cite{yosinski2015understanding} proposes two tools for visualizing and understanding what computations and neuron activations occur in the intermediate layers of deep neural networks (DNN).  Work by Schwartz-Ziv and Tishby suggest using a Information Plane visualization which captures the Mutual Information values that each layer preserves regarding the input and output variables of DNNs \cite{shwartz2017opening}. 

Other visualizations exist for other machine learning algorithms and learning tasks. Similar to saliency maps, sensitivity analysis provides a visualization that connects the inputs and outputs of $L$ \cite{saltelli2002sensitivity}.  Moreover, sensitivity analysis maps have been applied to tasks beyond image classification and to other black box machine learning algorithms such as ensemble trees \cite{cortez2013using}. For example, Coretz and Embrechts present five sensitivity analysis methods appropriate for both classification and regression tasks \cite{cortez2013using}. Zhang and Wallace present a sensitivity analysis for convolutional neural networks used in text classification \cite{zhang2015sensitivity}. 

Partial Dependency Plots (PDP) help visualize the average partial relationship between the predicted response of $L$ and one or more features within $F$ \cite{friedman2001greedy,goldstein2015peeking}. PDPs use feature analysis as a critical part of their interpretation, and are much more faithful and explicit than many of the other visualizations approaches in this section. However, as the primary output and interpretation tool is visual \cite{friedman2001greedy}, we have categorized it in this section. Examples include work by Hooker that uses ANOVA decomposition to help create this a Variable Interaction Network (VIN) visualization \cite{hooker2004discovering} and work by Goldstein et al. that extend the more classic PDP model by graphing the functional relationship between the predicted response and the feature for individual observations, thus making this a localized visualization \cite {goldstein2015peeking}. Similarly, Krause et al. provide a localized visualization to create partial dependence bars, a color bar representation of a PDP \cite{krause2016interacting}.

\subsection{Generating Explanations from Prototyping the Dataset's Input as Examples}
\label{What-protyping}
Similar to visualization tools, prototype selection also seeks to clarify the link between $L$'s input and output. However, while visualization tools focus on the input from $F$, prototyping focuses on $R$, seeking the existence of a subset of records similar to record, $r \in R$, being classified. This subset is meant to serve as an implicit explanation as to the correctness of the model as prototyping aims to find the minimal subset of input records that can serve as a distillation or condensed view of the dataset \cite{bien2011prototype}. 

Prototypes have been shown to help people better understand $L$'s decisions.  For example, work by Henricks et al. focuses on providing visual explanations for images that include class-discriminate information about other images that share common characteristics with the image being classified \cite{HendricksARDSD16}. The assumption here is that the information about similar pictures in the same class helps people better understand the decision of the algorithm. Bien and Tibshirani propose two methods for generating prototypes-- a LP relaxation with randomized rounding and a greedy approach \cite{bien2011prototype}. Work by Kim et al. suggested using maximum mean
discrepancy to generate prototypes \cite{kim2016examples}. In other work by Kim et al., they suggest using  a Bayesian Case Model (BCM) to generate prototypes \cite{kim2014bayesian}.

\subsection{Comparing the Six Basic Approaches for Generating Interpretations}
Referring back to Figure \ref{fig::Explicit-Faithful}, each of these approaches will differ along the axis of their level of explicitness and faithfulness. It has been previously noted that many of the visualization approaches produce interpretations that are not easily understood by people without an expert-level understanding of the problem being solved \cite{post} making them not very explicit. As they often provide justification and no direct interpretation of the logic in $L$, they are also not very faithful. As prototypes provide examples of similar classifications, they are often more explicit than visualizations as regular users can more easily understand their meaning. However, as they also do not attempt to directly explain $L$'s logic, they are not more faithful.   Other approaches, such as transparent ones, have high levels of both explicitness and faithfulness, but are typically limited to white box methods that facilitate these types of interpretability. Model and outcome tool approaches can potentially be geared to any user, making them very explicit, but are less faithful as the logic generated in $\I$ is not necessarily the same as that in $L$. When taken in combination with a white box algorithm, feature analysis methods can be very explicit and faithful. At times, they are used independently of $L$, potentially making them less faithful.

Referring back to Figure \ref{Figure1}, each of the approaches described in this section are labeled with the term within the explainability model described in Section \ref{definitions}. However, note the overlaps within the Venn Diagram as overlaps do exist between some of the approaches described in this section. While transparent approaches do link $\I$ and $L$, sometimes the link between these two elements is strengthened and/or described through an analysis of the $F$ commonly seen in Feature Analysis approaches. For example, the GAM and GA$^2$M approaches \cite{Caruana2015,LouCG12} use univariate and pairwise feature analysis methods respectively in their transparent models. While model outcome models such as SP-LIME pride themselves on being agnostic, e.g. no direct connection be assumed between $\I$ and $L$, they do use elements of feature analysis and visualization in creating their global interpretation of $L$ \cite{Ribeiro2016}. Similarly, the outcome explanation model, LIME, also uses feature analysis and visualization in creating its local interpretations of $L$ \cite{Ribeiro2016} for an instance of $r \in R$. Saliency maps are visualization that is based on identifying the features used for classifying a given picture \cite{fong2017interpretable} showing the potential overlap between visualization methods and feature analysis. However, at times, the identified salient features are used to create a outcome interpretation, as is the case in other work \cite{xu2015show}.  Similarly, work by Lei, Barzilay and Jaakkola generated visualizations of outcomes through analyzing which features were most useful to the model, again showing the intersection of these three approaches. Last, some prototype analysis tools, such as work by Henricks et al. use visual methods \cite{HendricksARDSD16}. Thus, we stress that the different types of interpretation approaches are often complementary and not mutually exclusive.

Given these differences of the explicitness and faithfulness of each of these approaches, it seems logical that the type of interface used for disseminated the system's interpretation will likely depend upon the level of the user's expertise and the type of interpretation that was generated. The idea of adaptable interfaces based on people's expertise was previously noted \cite{grudin1989case,shneiderman2002promoting,SteinGNRJ17}. In these systems, the type of information presented in the interface depends on the user's level of expertise. Accordingly, an interface might consider different types of interpretation or interpretation algorithms based on \emph{who} the end-user will be. Even among experts, it is reasonable to assume that different users will need different types of information. The different backgrounds of legal experts, scientists, safety engineers, and researchers may necessitate different types of interfaces \cite{doshi2017towards}.

\section{\emph{When} Should Information be Presented?}
\label{when}

Explanations can be categorized based on \emph{when} the interpretation is presented:
\begin{enumerate}
\item Before the task  
\item Continuing explanations throughout the task  
\item After the task  
\end{enumerate}

Some agents may present their interpretation before the task is executed as either justification \cite{biran2017explanation}, conceptualization or proof of fairness of an agent's intended  action \cite{dwork2012fairness}. Other agents may present their explanation during task execution, especially if this information is important to explain when the agent fails so it will be trusted to correct the error \cite{jennings2014human,abs-1806-00069,Guidotti2018}. Other agents provide explanations after actions are carried out \cite{langley2017explainable}, to be used for retrospective reports \cite{Lipton16a}. 

It is important to note that not all approaches for \emph{what} can be generated, as per Section \ref{What} support all of these possibilities. While all methods can be used for analysis after the task, many of these methods use post-hoc analysis that separates $L$ from $\I$. Thus, if fairness needs to be checked before task execution, the lack of connection between $L$ from $\I$ in model and outcome explanations, visualizations, and prototypes make this difficult to accurately check. Transparent methods could fulfill this requirement due to their inherent faithfulness. Feature analysis methods including, but not limited to GAM, GA$^2$M, and PDP \cite{Caruana2015,friedman2001greedy,goldstein2015peeking,lou2013accurate} can check the connection between inputs and outputs, thus confirming fairness or other legal requirements are met even before task execution. 

The choice of \emph{when} to present the explanation is not exclusive. Agents might supply various explanations at various times, before, during and after the task is carried out. Building on the taxonomy in Section \ref{Why}, if explainability is critical for the system to begin functioning, then it stands to reason that this knowledge must be presented at the beginning of the task, thus enabling the user to determine whether to accept the agent's recommendation \cite{sheh2017did}. However, if it is beneficial to build trust / user acceptance, then it might be directed during the task, especially if the agent erred.  If the purpose of the explanation is to justify the agent's choice from a legal perspective then we may need to certify that decision before the agent acts (preventative) or after the act (accusatory). But, if the goal is conceptualization, especially in the form of knowledge discovery and/or to support future decisions, then the need for explanation after task execution is equally critical. These possibilities are not inherently mutually exclusive. For example, work by  Vigano and Magazzeni \cite{vigano2018explainable} claims that explanations should be provided throughout all stages of the systems lifecycle within security systems. They describe how explanations should begin as the system is designed an implemented, continue through use, analysis and change and maybe even when it is replaced. One may argue whether this is crucial for all systems or only for security systems that are discussed in their work, but it is surely a point to consider.

\section{\emph{How} can Explanations be Evaluated?} \label{How}
It was previously noted that little agreement currently exists about how to define explainability and interpretability which may be adding to the difficulty in properly evaluating it \cite{doshi2017towards}.  In order to address this point, we first clearly defined these terms in Section \ref{definitions}, and then proceeded to consider questions of \emph{why}, \emph{what}, \emph{when} and \emph{how} based on these definitions. 

As we discuss in this section, creating a general evaluation framework is still an open challenge as these issues are often intrinsically connected. For example, the detail of an explanation is often dependent on \emph{why} that explanation is needed.  An expert will likely differ from a regular user regarding \emph{why} an explanation is needed, will often need these explanations at different times, e.g. before or after the task (\emph{when}), and may require different types of explanations and  interfaces (\emph{what} and \emph{how}). At other times multiple facets of explanation exist even within one category.  A DSS system is built to support a user's decision, thus making explainability a critical issue.  However, these systems will still likely benefit from better explanations, so that the user trusts those explanations. Similarly, a scientist pursuing knowledge discovery may need to analyze and interact with information presented before, during and after a task's completion (\emph{when}).  Thus, multiple goals must often be considered and evaluated.

To date, there is little consensus about how to quantify these interconnections. Many works evaluated explainability as a binary value-- either it works or it doesn't. Within these papers, an explanation is inspected to insure that it provides the necessary information  in the context of a specific application \cite{lei2016rationalizing,Ribeiro2016}. If it does so, it is judged as a success, even if other approaches may have been more successful. 
To help quantify evaluations, Doshi-Velez and Kim suggested a taxonomy of three types of evaluation tasks that can be used for evaluation: application, human, and functionally grounded \cite{doshi2017towards}. In their model, application grounded tasks are meant for experts attempting to execute a task and the evaluation focuses on how well the task was completed. Human-grounded tasks are simplified and can be performed with regular-users. They conceded that it is not clear what the evaluation goal of this task need be but recommended simplified evaluation metrics, such as reporting which explanation they preferred. Work by Mohseni and Ragan suggested creating canonical datasets of this type that could quantify differences between interpretable algorithms \cite{mohseni2018human}. They proposed annotating regions in images, and words in texts that provide an explanation. The output of any new interpretation algorithm, $\I$, could be compared to the user annotations that provide a ground-truth. This approach is still goal-oriented and thus they classify their task as a human-grounded task (e.g. having $\I$ match the human explanation). Doshi-Velez and Kim's last type of evaluation task is functionality-grounded where some objective criteria for evaluation is predefined (e.g. the ratio of decision tree size to model accuracy). The main advantage to this type of evaluation is the evaluation of $\I$ can be quantified without any need for user studies.

This taxonomy provides three important types of tasks that can be used in evaluate explainability and interpretability, but these researchers do not propose how to quantify the effectiveness of the key components within $\E$ and $\I$. This paper's main point is that questions surrounding the system's need for \emph{why}, \emph{what}, \emph{when} and \emph{how} about explainability must be addressed. These elements can and should be quantified, while also considering trade-offs between these categories as well as elements within $\I$. Issues about algorithms' explicitness, faithfulness and transparency must be explicitly evaluated while balancing the agent's and user's performance requirements, including the agent's fairness and prediction accuracy, and the user's performance and acceptance of $\E$'s.

Given this, we suggest explicitly evaluating the following three elements in Human-Agent Systems: The quantifiable performance of the agent's learning, $L$, its level of interpretation, $\I$, and human's understanding, $\E$. For example, in a movie recommendation system the three scores would be described as follows: The  score of for $L$ is based on standard metrics for evaluating recommendation predictions (i.e. accuracy, precision and/or recall). A score can also be given to $\I$ that reflects how much explicitness, faithfulness and transparency exist in $\I$ according to objective criteria described below. The score for $\E$ should be quantified based on the user's performance. As the goal of the system is to yield predictions that the user understands so they will be accepted, we should quantify the impact of $\I$ on the user's behavior. Thus, we suggest an evaluation score that quantifies:

\begin{enumerate}
\item A score for $L$, the performance of the agent's prediction 
\item A score for $\I$, the interpretation given to the user 
\item A score for $\E$, the user's acceptance of $\I$
\end{enumerate}

As described in previous sections, a complex interplay exists between these three elements.  There is often a trade-off between the performance of $L$ and the explicitness of $\I$ that can be produced from $L$ (see Figure \ref{fig::Explicit-Predict}). White-box algorithms are more explicit and can even be transparent, but typically have lower performance. Higher accuracy algorithms, such as neural networks, are typically less explicit and faithful (see Figure \ref{fig::Explicit-Faithful}). Thus, agents with lower performance scores for $L$ will likely have higher scores for $\E$, especially if explicitness and faithfulness are important and quantifiable within the system. Furthermore, different user types and interfaces will be effected by the type of agent design and a total measure is needed to weigh all parameters that are needed by a system into account. For example, an agent that was designed to support an expert user is different from one provided to a regular user.  

Another equally important element of the system is how well the person executed her system task(s) given $\I$. In theory, multiple goals for $\E$ may exist for the human user such as immediate performance vs. long-term knowledge acquisition. These may be complementary or in conflict.  For example, assume the explanation goal of a system is to support a person's ability to purchase items in a time-constrained environment (e.g. online stock purchasing).  The greater detail contained within the agent's explanations on one hand instill improved confidence within the user, but also will take more time to read and process, which may prevent the user from capitalizing on certain quickly passing market fluctuations. Thus, some measure should likely be introduced to reason about different goals for the explanation and the relative strengths of various explanations, their interfaces, and the algorithms that generate those explanations.

To capture these properties, we propose an overall utility to quantify the complementary and contradictory goals for $L, \I$, and $\E$ as the weighted product:

\begin{equation}
Utility = \prod_{n=1}^{NumGoals} Imp_n*Grade_n
\label{equation-utility}
\end{equation}
\begin{equation}
\sum_{n=1}^{NumGoals} Imp_n = 1
\end{equation}

We define $NumGoals$ as the number of goals in an the system.  $L, \I$, and $\E$ each have an overall objective and the system meets this object through all of these goals.  The objective for $L$ is to provide predictions for $T$ using 
$R \times F$. The ability of the system to meet this objective is measured through machine learning performance metrics that quantify the goals of high accuracy, recall, precision, F-measure, mean average precision, and mean squared error. A goal for $L$ can also be that $L$ exhibits fairness, which often is a hard-constraint due to legal considerations. The objective for $\I$ is to provide a representation of $L$'s logic that is understandable to the user. This success of this objective can be measured by goals for $\I$ to have the highest levels of explicitness, faithfulness, and transparency. Other papers have suggested additional goals for $\I$ including justification \cite{kofod2008explanatory} and completeness \cite{abs-1806-00069} that we argue are included in goals of explicitness and faithfulness respectively. The objective for $\E$ is that the person will understand  $L$ using $\I$. This can be measured through the goal that the user's performance be improved given $\I$. Additional goals include those specified in Section \ref{Why} including guaranteeing safety concerns, trust, and knowledge / scientific discovery. Goals such as to the timing of when interpretations were present (e.g. ``presenting during task as required") are likely hard-constraints (e.g. either it was done at the correct time or not). $Imp_n$ is the importance weight we give to the $n_{th}$ goal such that $0<Imp_n<=1$.  Similarly, $Grade_n$ is the score we give to  the $n_{th}$ goal, and we require that $0<=Grade_n<=1$.

While this model helps quantify the interplay of multiple explanation goals, either inherently complimentary or contradictory, a fundamental question lingers about how to set the values of $NumGoals$, $Imp_n$ and $Grade_n$.  Of these values, we argue that $NumGoals$ is the easiest to define and should be clearly defined in advance, by addressing the key elements of explainability as defined in Section \ref{definitions} (e.g. fairness, transparency, explicitness, etc.) as well as the questions about \emph{why}, \emph{what}, \emph{when} and \emph{how} discussed in subsequent sections. Many of these goals are hard-constraints that must be fulfilled. For example, assuming a system must exhibit fairness, but fails to do so, the grade for this goal will be zero. As the overall utility is the product of all goals and their grades, the net utility of the system will be zero as the hard-constraint was not met. 

However, $Imp_n$ and $Grade_n$ are much more difficult to quantify in many real-world applications, the first category within the taxonomy of Doshi-Velez and Kim \cite{doshi2017towards}. We assume that either users themselves, the system's designer, or outside third party organizations (e.g. governments) can quantify these values including the trade-offs between them.  For example, a system designer may determine that a transparent system is necessary or desirable through setting $Imp_n$ for this goal. (Assuming transparency is not needed at all, it should be removed as to avoid a zero grade making the net utility be zero.) If a high level of transparency can only be obtained at the cost of a lower accuracy, two conflicting goals exist and the system designer will need to decide the relative importance of both goals, i.e. $Imp_n$ for each goal, so the optimal trade-off can be found. 

Further complicating the evaluation calculation, to date no quantifiable measurements exists for the goals in $\I$.  In contrast, very quantifiable metrics exist for $L$ and to a lesser degree, for those in $\E$. $L$ is relatively easily quantifiable through accepted measures such as accuracy, precision, and recall. We suggest that goals within $\E$ be measured through known tools for quantifying the user experience as is typically done in HCI studies. In addition to grading the user's performance, accepted user performance metrics such as the NASA-TLX (Task Load Index)
\cite{hart1988development} and the System Usability Scale \cite{brooke1996sus} can measure these goals by focusing and the user's mental workload \cite{rubio2004evaluation}. Specific to measuring explanations, it has been suggested that simplicity and satisfaction be considered as a potential metrics \cite{gelderman1998relation,lombrozo2007simplicity}. However, to date, no accepted measures exist for quantifying the elements of $\I$ described in this paper. For example, what should the scale for grading an algorithm's explicitness or faithfulness be?  Should they be normalized between 0 and 1?  If so, on what basis? This is an open challenge that we believe needs to be addressed in the future. 

We agree with previous work that simplified tasks, functional metrics and binary grades should all be used be to help tractably evaluate explainable systems \cite{doshi2017towards,lei2016rationalizing,Ribeiro2016}. Examples of simplified tasks include the creation of canonical datasets where an objective truth exists about correct interpretations and explanations \cite{mohseni2018human}.  A second approach is to quantify the relative value of different approaches and only give a non-zero score to the one that they feel is best \cite{lei2016rationalizing,Ribeiro2016}.  Thus, the scoring function $Grade_n$ could be made boolean (e.g. either the interpretation is either explicit or it isn't), greatly simplifying calculating the system's total value. or through using binary evaluations. In both cases, questions about how to set $Grade_n$ can be resolved in this way. A third set of approaches suggests quantifying elements of $\I$ through functional metrics about the interpretation \cite{doshi2017towards}. The assumption behind this approach is that as the size of machine learning models grow, they are less explicit. Thus, models with large numbers of nodes / hidden layers (e.g. in deep neural networks), parameter values (for regression and SVM models), the number of rules (rule-based models), or the depth (in decision trees) are less preferable to those with fewer numbers of these values \cite{Explain2018}. Following Section \ref{What}, we could also create an objective metric based on the number of features generated from feature selection that are used to create the model. The advantage to both of these approaches is the value of explanation's $Grade_n$ can be set independently of the system's specific task. A fourth  approach  is to create simplified accepted tasks or case studies, potentially where simulations of human behavior could be used for repeatedly for evaluation across different algorithms, interfaces, and approaches \cite{doshi2017towards,DoranSB17}. Similarly, this could create a standardization for all values of $NumGoals$, $Imp_n$ and $Grade_n$, again greatly aiding in the evaluation process.  Currently, no such canonical tasks have been universally accepted, leaving this issue as an open challenge.

\section{Discussion}
\label{discuss}

Explainability of Human-Agent Systems is a complex matter, involving multiple, and sometime contradicting, aspects. This paper is unique in the integrated approach we take to addressing these questions. In Section \ref{sec:definitions} we define key terms based on previous work \cite{abs-1806-07538,dwork2012fairness,Guidotti2018,4938655} combined with contributions of our own. We then use these terms to analyze extensively  \emph{why} these explanations are needed, if the recipient has a specific skill-level, what is the mechanism for generating these explanations, \emph{when} it should be presented, \emph{what} interpretations can be generated and \emph{how} the entire system can be evaluated.

To help focus the reader on previous contributions and how they relate to this work, Table \ref{mapping} presents a mapping of papers that we encountered to the main component of each of the fundamental questions regarding explainability we addressed. For each paper that we included in the mapping in Table \ref{mapping} we provide the citation number, and indicate the questions about explainability they address. Please note that there are no studies that touch on all aspects of explainability presented in this paper. However, we do agree that certain papers touch upon more than one of these topics. In this case, we categorized the paper as per which issue we felt was the paper's focus. As this is a dynamic field, new papers do exist and we do not claim that this list is exhaustive. However, we believe that all papers, both current and future, can be categorized as per the divisions in this table.

It is interesting to note the contrast between the number of papers aimed at clearly defining these terms in line 2, to those who that questions of \emph{why, who when and how} in lines 3, 4, 11, and 12 with the number of papers that address the question of \emph{what} in lines 5--10. Most papers are unfortunately geared to addressing only this issue without focusing on other key points about explainability. This paper's key point is that these other questions are actually extremely important questions to ask, as they heavily affect the question of \emph{what} explanation to generate.

\begin{table}[ht] 
\centering
\begin{tabular}{|p{0.6cm}|p{4cm}|p{8.1cm}|}\hline
\multicolumn{2}{|l|}{Section}& Papers \\\hline

\multicolumn{2}{|l|}{Definitions  (Sec. \ref{sec:definitions})}& \cite{achinstein1983nature},\cite{chen2014situation},\cite{doshi2017towards},\cite{gregor1999explanations},\cite{Guidotti2018},\cite{schank1986explanation},\cite{sormo2005explanation},\cite{van198511} \\\hline

 \multicolumn{2}{|l|}{Why (Sec. \ref{Why}) }& \cite{doshi2017towards}, \cite{XAIP2017},\cite{Guidotti2018},\cite{Ribeiro2016},\cite{vigano2018explainable}\\\hline
 
\multicolumn{2}{|l|}{ Who (Sec. \ref{Who})} &  \cite{doshi2017towards},\cite{HendricksARDSD16},\cite{Simonyan2013DeepIC},\cite{vigano2018explainable} \\\hline
 
\multirow{6}{*}{What}  & Transparent (Sec. \ref{What-transparent}) & \cite{bagley2001logistic},\cite{bellazzi2008predictive},\cite{zupan2000machine},\cite{Caruana2015},\cite{cheng1999comparing},\cite{clark1989cn2},\cite{Explain2018},\cite{dreiseitl2002logistic},\cite{Freitas2014},\cite{gunning2017explainable},\cite{hara2016making},\cite{katafigiotis2018stone},
\cite{kim2014bayesian},\cite{kononenko1993inductive},\cite{last2004compact},\cite{lavravc1999selected},\cite{LouCG12},\cite{lou2013accurate},\cite{michalski1999learning},\cite{mood2010logistic},\cite{quinlan1986induction},\cite{sormo2005explanation}
 \\\cline{2-3}
  &Feature analysis (Sec. \ref{What-feature analysis}) & \cite{Kononenko99},\cite{vellido2012robust},\cite{vellido2012making} \\\cline{2-3}
  & Model tool (Sec. \ref{What-Model outcome}) & \cite{arbatli1997rule},\cite{Augasta2012},\cite{Boz2002},\cite{Chipman_makingsense},\cite{craven1994using},\cite{Craven1995},\cite{deng2014interpreting},\cite{Domingos1998},\cite{frosst2017distilling},\cite{hara2016making},\cite{4938655},\cite{Kahramanli},
  \cite{KRISHNAN1999},\cite{Mohamed2011},\cite{Ribeiro2016},\cite{4167900},\cite{tan2016tree},\cite{tolomei2017interpretable},\cite{whitmore2018explicating},\cite{zhou2016interpreting},\cite{zhou2003extracting}\\\cline{2-3}
  & Outcome tool (Sec. \ref{What-outcome explanation}) & \cite{Baehrens2010},\cite{lundberg2017unified},\cite{miller2017explanation}\\\cline{2-3}
  & Visualization (Sec. \ref{What-visualization}) & \cite{cortez2013using},\cite{fong2017interpretable},\cite{goldstein2015peeking},\cite{guo2010novel},\cite{hooker2004discovering},\cite{hu2018explainable},\cite{kim2018textual},\cite{krause2016interacting},\cite{lei2016rationalizing},\cite{saltelli2002sensitivity},\cite{selvaraju2017grad},
  \cite{shwartz2017opening},\cite{Simonyan2013DeepIC},\cite{xu2015show},\cite{yosinski2015understanding},\cite{zhang2015sensitivity},\cite{zhou2016learning}\\\cline{2-3}
  & Prototyping (Sec. \ref{What-protyping}) & \cite{bien2011prototype},\cite{HendricksARDSD16},\cite{kim2016examples},\cite{kim2014bayesian}\\\hline
 
 \multicolumn{2}{|l|}{When (Sec. \ref{when}) }&  \cite{langley2017explainable},\cite{Lipton16a},\cite{sheh2017did},\cite{vigano2018explainable} \\\hline
 
 \multicolumn{2}{|l|}{How (Sec. \ref{How}) }&  \cite{brooke1996sus},\cite{Explain2018},\cite{doshi2017towards},\cite{hart1988development},\cite{mohseni2018human}\\\hline

\end{tabular}
\caption{Each section in our paper discusses an aspect of explainability. We list the papers discussed for each of these aspects.}
\label{mapping}
\end{table}

To date, several excellent surveys exist on this topic, each of which addresses aspects of the taxonomy and evaluation framework about the \emph{Why, Who, What, When} and \emph{How} for this issue \cite{CHI2018,doshi2017towards,dreiseitl2002logistic,Guidotti2018,post,zhang2018visual}. In this paper, we claim all issues regarding  explainability are interwoven. This leads to a  different analysis of explainability. For example, two recent  papers \cite{Guidotti2018,post} provide excellent coverage of the topic of interpretability of black box models. Another survey is even more specific, focusing on visualization tools for deep neural networks \cite{zhang2018visual}. Conversely, a different survey \cite{dreiseitl2002logistic} focuses on white box algorithms. Another survey \cite{CHI2018} is written from an HCI perspective with the aim of using topic model to undercover trends within the HCI perspective of explainable systems. A different one focuses on evaluation possibilities \cite{doshi2017towards}. We encourage the reader to study these papers in conjunction with our work as they address specific points within this paper. 

We believe it is a mistake to exclusively focus on either white box or black box algorithms as the source of interpretability within explainable systems. While at many times black box models perform better, this metric is only one aspect of the evaluation, the agent's performance. If the motivation for requiring explainability rises from legal issues, only transparency may fulfill this need as it  allows for levels of explicitness and faithfulness not existent within black box methods. Assuming the goal for explainability is to guarantee safety concerns as per Section \ref{Why}, then this may be a hard-constraint which precludes other methods for generating explanations. If the user is not an expert (as per Section \ref{Who}), then black box visualization tools are likely not useful as they are typically only readily understood by experts \cite{post}.  However, if performance is a larger concern, and the goal of explainability is to build trust, then explanations built upon prototyping as per Section \ref{What} may be sufficient for a regular user, despite their not being explicit or faithful.   If the target of explanation is an external legal entity concerned with the algorithm's faithfulness, then feature analysis is likely sufficient to assuage concerns that a specific set of features are not being abused. In all cases, the evaluation of the algorithm will likely change as the type of user, timing, and type of explanation being generated will likely need to be fundamentally different giving the large variations between the described elements within the explanation.

Looking forward, we identify several open issues that result from our analysis of this issue:
\begin{enumerate}
\item What measurements for explicitness and faithfulness can be created for a given algorithm $L$?
\item What canonical tasks can be developed for measuring $\I$ and $\E$?
\item What tasks can be identified where one method for \emph{what} type of interpretation (Section \ref{What}) is clearly better?
\item Can the level of interpretability within black box models equal that of white box ones?
\item Is justification ever advantageous over interpretable models with higher levels of explicitness and faithfulness?
\end{enumerate}

Previously, Doshi-Velez and Kim \cite{doshi2017towards} posed an open challenge to identify what important factors should be considered in evaluating interpretation and evaluation quality. We believe this paper has clearly identified explicitness and faithfulness as these key elements within $\I$. Nonetheless, further work is necessary to quantify these elements, especially as different algorithms described in Section \ref{What} of this paper differ in this regard, even within the six categories we presented. Similarly, even without quantifying these elements, canonical tasks are needed where all six types of these algorithms can be implemented to facilitate the relative performance of these algorithms in the performance goals of $L$, $\I$, and $\E$. We hope that certain tasks could be identified where one type of algorithm is clearly better in addressing one aspect of the issues we raise, such as trust.  While many works exclusively focus on black box interpretation due to the high performance of these agents, it is not clear if these agents are suited for all situations and if the level of interpretability of these algorithms will ever reach that of white box agents. Until this happens their suitability must be questioned for certain situations such as fulfilling legal requirements for explainability or safety. Last, we have intentionally limited our discussion about justification as there is no need for these models to be part of an explainable system. Nonetheless, we do believe that justification is important for many of the goals of explainable systems. One open question is to further study the relationship between justification and explainability such that tasks could be identified when one approach is advantageous over the other, or if hybrid methods that incorporate elements of both approaches should be used.

\section{Conclusion} \label{conclusion}
We presented a framework designed to enable comparison and evaluation of explainability  in Human-Agent Systems. As Human-Agent Systems are diverse and complex, there is no ``one explanation type fits all". Each agent must have its requirements and goals mapped out, and the appropriate explanation chosen. We focused on agents that use machine learning and provided an attempt to define this new field. 

Our first contribution is a proposed clear and consistent set of definitions for the key terms of explainability, interpretability, transparency, fairness, explicitness and faithfulness in learning algorithms that interact with people. Using these definitions, we systematically address  five questions about explainability: \emph{Why, Who, What, When} and \emph{How} can be answered. These questions define the various aspects of the explanation for the system. In designing an agent one must first establish \emph{why} the system requires explanation, as this will affect the answer to the other questions. Next, one must determine \emph{who} is the target of this explanation, \emph{what}  type of explanation is needed and \emph{when} it must be presented. Finally, the question of \emph{how} to evaluate the explanation must be addressed. For each of the questions we presented possible approaches, and discussed when each possibility is likely most appropriate. Various factors affect the answers to these questions. We discussed how the degree of control of the user over the agent affects the need for explainability.  We investigated how  different user types might affect the explanation. We also discussed how the type of learning that agents perform will affect the explanation that is provided. We then discussed parameters for \emph{when} to present the information.  Finally we presented an evaluation measure, composed of three elements for comparing systems. Our proposed utility is capable of combining all the aspects of the system: the machine learning algorithm, user performance and the explanation, into a single measure. We discussed the strengths and  limitations of our proposed measure. While the measure provides a means for comparison, its main limitation relates to the elements that can potentially be subjective in determining the values of the parameters. 

We hope that the definitions presented in this paper will serve as a basis for future studies about the five questions about explainability that we present, particularly in the proper evaluation of explainability in Human-Agent Systems. Furthermore, we hope additional researchers will use this framework for further analysis of new algorithms, including suggesting extensions, in this emerging field. Towards this goal, we identified several open issues based on the analysis presented in this paper.

\bibliographystyle{plain}
\bibliography{references}

\begin{thebibliography}{100}

\bibitem{CHI2018}
Ashraf Abdul, Jo~Vermeulen, Danding Wang, Brian~Y. Lim, and Mohan Kankanhalli.
\newblock Trends and trajectories for explainable, accountable and intelligible
  systems: An hci research agenda.
\newblock In {\em Proceedings of the 2018 CHI Conference on Human Factors in
  Computing Systems}, pages 582:1--582:18, 2018.

\bibitem{achinstein1983nature}
Peter Achinstein.
\newblock {\em The nature of explanation}.
\newblock Oxford University Press, 1983.

\bibitem{Adam2008}
Fr{\'e}d{\'e}ric Adam.
\newblock {\em Encyclopedia of decision making and decision support
  technologies}, volume~2.
\newblock IGI Global, 2008.

\bibitem{ante-2018}
M.~A. {Ahmad}, A.~{Teredesai}, and C.~{Eckert}.
\newblock Interpretable machine learning in healthcare.
\newblock In {\em 2018 IEEE International Conference on Healthcare Informatics
  (ICHI)}, pages 447--447, 2018.

\bibitem{abs-1806-07538}
David Alvarez{-}Melis and Tommi~S. Jaakkola.
\newblock Towards robust interpretability with self-explaining neural networks.
\newblock {\em CoRR}, abs/1806.07538, 2018.

\bibitem{amir2013plan}
Ofra Amir and Kobi Gal.
\newblock Plan recognition and visualization in exploratory learning
  environments.
\newblock {\em ACM Transactions on Interactive Intelligent Systems (TiiS)},
  3(3):16, 2013.

\bibitem{arbatli1997rule}
A~Duygu Arbatli and H~Levent Akin.
\newblock Rule extraction from trained neural networks using genetic
  algorithms.
\newblock {\em Nonlinear Analysis: Theory, Methods \& Applications},
  30(3):1639--1648, 1997.

\bibitem{Augasta2012}
M.~Gethsiyal Augasta and T.~Kathirvalavakumar.
\newblock Reverse engineering the neural networks for rule extraction in
  classification problems.
\newblock {\em Neural Process. Lett.}, 35(2):131--150, April 2012.

\bibitem{azaria2015strategic}
Amos Azaria, Zinovi Rabinovich, Claudia~V Goldman, and Sarit Kraus.
\newblock Strategic information disclosure to people with multiple
  alternatives.
\newblock {\em ACM Transactions on Intelligent Systems and Technology (TIST)},
  5(4):64, 2015.

\bibitem{azaria2014agent}
Amos Azaria, Ariella Richardson, and Sarit Kraus.
\newblock An agent for the prospect presentation problem.
\newblock In {\em Proceedings of the 2014 international conference on
  Autonomous agents and multi-agent systems}, pages 989--996. International
  Foundation for Autonomous Agents and Multiagent Systems, 2014.

\bibitem{Baehrens2010}
David Baehrens, Timon Schroeter, Stefan Harmeling, Motoaki Kawanabe, Katja
  Hansen, and Klaus-Robert M\"{u}ller.
\newblock How to explain individual classification decisions.
\newblock {\em J. Mach. Learn. Res.}, 11:1803--1831, August 2010.

\bibitem{bagley2001logistic}
Steven~C Bagley, Halbert White, and Beatrice~A Golomb.
\newblock Logistic regression in the medical literature:: Standards for use and
  reporting, with particular attention to one medical domain.
\newblock {\em Journal of clinical epidemiology}, 54(10):979--985, 2001.

\bibitem{barrett2017making}
Samuel Barrett, Avi Rosenfeld, Sarit Kraus, and Peter Stone.
\newblock Making friends on the fly: Cooperating with new teammates.
\newblock {\em Artificial Intelligence}, 242:132--171, 2017.

\bibitem{bellazzi2008predictive}
Riccardo Bellazzi and Blaz Zupan.
\newblock Predictive data mining in clinical medicine: current issues and
  guidelines.
\newblock {\em International journal of medical informatics}, 77(2):81--97,
  2008.

\bibitem{bien2011prototype}
Jacob Bien and Robert Tibshirani.
\newblock Prototype selection for interpretable classification.
\newblock {\em The Annals of Applied Statistics}, pages 2403--2424, 2011.

\bibitem{biran2017explanation}
Or~Biran and Courtenay Cotton.
\newblock Explanation and justification in machine learning: A survey.
\newblock In {\em IJCAI-17 Workshop on Explainable AI (XAI)}, 2017.

\bibitem{zupan2000machine}
Michael W Kattan J Robert Beck Ivan~Bratko Blaz~Zupan, Janez~Demsar.
\newblock Machine learning for survival analysis: a case study on recurrence of
  prostate cancer.
\newblock {\em Artificial intelligence in medicine}, 20(1):59--75, 2000.

\bibitem{Boz2002}
Olcay Boz.
\newblock Extracting decision trees from trained neural networks.
\newblock In {\em Proceedings of the Eighth ACM SIGKDD International Conference
  on Knowledge Discovery and Data Mining}, pages 456--461, 2002.

\bibitem{brooke1996sus}
John Brooke et~al.
\newblock Sus-a quick and dirty usability scale.
\newblock {\em Usability evaluation in industry}, 189(194):4--7, 1996.

\bibitem{Caruana2015}
Rich Caruana, Yin Lou, Johannes Gehrke, Paul Koch, Marc Sturm, and Noemie
  Elhadad.
\newblock Intelligible models for healthcare: Predicting pneumonia risk and
  hospital 30-day readmission.
\newblock In {\em Proceedings of the 21th ACM SIGKDD International Conference
  on Knowledge Discovery and Data Mining}, pages 1721--1730, 2015.

\bibitem{chen2014situation}
Jessie~Y Chen, Katelyn Procci, Michael Boyce, Julia Wright, Andre Garcia, and
  Michael Barnes.
\newblock Situation awareness-based agent transparency.
\newblock Technical report, Army Research Lab Aberdeen Proving Ground MD Human
  Research and Engineering Directorate, 2014.

\bibitem{cheng1999comparing}
Jie Cheng and Russell Greiner.
\newblock Comparing bayesian network classifiers.
\newblock In {\em Proceedings of the Fifteenth conference on Uncertainty in
  artificial intelligence}, pages 101--108, 1999.

\bibitem{Chipman_makingsense}
H.~A. Chipman, E.~I. George, and R.~E. Mcculloch.
\newblock Making sense of a forest of trees.
\newblock In {\em Proceedings of the 30th Symposium on the Interface}, pages
  84--92, 1998.

\bibitem{clancey1983epistemology}
William~J Clancey.
\newblock The epistemology of a rule-based expert system—a framework for
  explanation.
\newblock {\em Artificial intelligence}, 20(3):215--251, 1983.

\bibitem{clancey1982neomycin}
William~J Clancey and Reed Letsinger.
\newblock {\em NEOMYCIN: Reconfiguring a rule-based expert system for
  application to teaching}.
\newblock Department of Computer Science, Stanford University, 1982.

\bibitem{clark1989cn2}
Peter Clark and Tim Niblett.
\newblock The cn2 induction algorithm.
\newblock {\em Machine learning}, 3(4):261--283, 1989.

\bibitem{corchado2003constructing}
Juan~M Corchado and Rosal{\'\i}a Laza.
\newblock Constructing deliberative agents with case-based reasoning
  technology.
\newblock {\em International Journal of Intelligent Systems},
  18(12):1227--1241, 2003.

\bibitem{cortez2013using}
Paulo Cortez and Mark~J Embrechts.
\newblock Using sensitivity analysis and visualization techniques to open black
  box data mining models.
\newblock {\em Information Sciences}, 225:1--17, 2013.

\bibitem{cox2011metareasoning}
Michael~T Cox and Anita Raja.
\newblock {\em Metareasoning: Thinking about thinking}.
\newblock MIT Press, 2011.

\bibitem{craven1994using}
Mark~W Craven and Jude~W Shavlik.
\newblock Using sampling and queries to extract rules from trained neural
  networks.
\newblock In {\em Machine Learning Proceedings 1994}, pages 37--45. 1994.

\bibitem{Craven1995}
Mark~W. Craven and Jude~W. Shavlik.
\newblock Extracting tree-structured representations of trained networks.
\newblock In {\em Proceedings of the 8th International Conference on Neural
  Information Processing Systems}, NIPS'95, pages 24--30, Cambridge, MA, USA,
  1995. MIT Press.

\bibitem{crockett2016data}
David Crockett and Brian Eliason.
\newblock What is data mining in healthcare?, 2016.

\bibitem{Explain2018}
Hoa~Khanh Dam, Truyen Tran, and Aditya Ghose.
\newblock Explainable software analytics.
\newblock {\em CoRR}, abs/1802.00603, 2018.

\bibitem{deng2014interpreting}
Houtao Deng.
\newblock Interpreting tree ensembles with intrees.
\newblock {\em arXiv preprint arXiv:1408.5456}, 2014.

\bibitem{Domingos1998}
Pedro Domingos.
\newblock Knowledge discovery via multiple models.
\newblock {\em Intell. Data Anal.}, 2(3):187--202, May 1998.

\bibitem{DoranSB17}
Derek Doran, Sarah Schulz, and Tarek~R. Besold.
\newblock What does explainable {AI} really mean? {A} new conceptualization of
  perspectives.
\newblock In {\em Proceedings of the First International Workshop on
  Comprehensibility and Explanation in {AI} and {ML}}, 2017.

\bibitem{doshi2017towards}
Finale Doshi-Velez and Been Kim.
\newblock Towards a rigorous science of interpretable machine learning.
\newblock {\em arXiv preprint arXiv:1702.08608}, 2017.

\bibitem{dreiseitl2002logistic}
Stephan Dreiseitl and Lucila Ohno-Machado.
\newblock Logistic regression and artificial neural network classification
  models: a methodology review.
\newblock {\em Journal of biomedical informatics}, 35(5-6):352--359, 2002.

\bibitem{dwork2012fairness}
Cynthia Dwork, Moritz Hardt, Toniann Pitassi, Omer Reingold, and Richard Zemel.
\newblock Fairness through awareness.
\newblock In {\em Proceedings of the 3rd innovations in theoretical computer
  science conference}, pages 214--226, 2012.

\bibitem{fong2017interpretable}
Ruth~C Fong and Andrea Vedaldi.
\newblock Interpretable explanations of black boxes by meaningful perturbation.
\newblock In {\em 2017 IEEE international conference on computer vision
  (ICCV)}, pages 3449--3457, 2017.

\bibitem{XAIP2017}
Maria Fox, Derek Long, and Daniele Magazzeni.
\newblock Explainable planning.
\newblock {\em CoRR}, abs/1709.10256, 2017.

\bibitem{Freitas2014}
Alex~A. Freitas.
\newblock Comprehensible classification models: A position paper.
\newblock {\em SIGKDD Explor. Newsl.}, 15(1):1--10, March 2014.

\bibitem{friedman2001greedy}
Jerome~H Friedman.
\newblock Greedy function approximation: a gradient boosting machine.
\newblock {\em Annals of statistics}, pages 1189--1232, 2001.

\bibitem{frosst2017distilling}
Nicholas Frosst and Geoffrey Hinton.
\newblock Distilling a neural network into a soft decision tree.
\newblock {\em arXiv preprint arXiv:1711.09784}, 2017.

\bibitem{Garfinkel2017}
Simson Garfinkel, Jeanna Matthews, Stuart~S. Shapiro, and Jonathan~M. Smith.
\newblock Toward algorithmic transparency and accountability.
\newblock {\em Commun. ACM}, 60(9):5--5, August 2017.

\bibitem{gelderman1998relation}
Maarten Gelderman.
\newblock The relation between user satisfaction, usage of information systems
  and performance.
\newblock {\em Information \& management}, 34(1):11--18, 1998.

\bibitem{gilbert1989explanation}
Nigel Gilbert.
\newblock Explanation and dialogue.
\newblock {\em The Knowledge Engineering Review}, 4(3):235--247, 1989.

\bibitem{abs-1806-00069}
Leilani~H. Gilpin, David Bau, Ben~Z. Yuan, Ayesha Bajwa, Michael Specter, and
  Lalana Kagal.
\newblock Explaining explanations: An approach to evaluating interpretability
  of machine learning.
\newblock {\em CoRR}, abs/1806.00069, 2018.

\bibitem{goldstein2015peeking}
Alex Goldstein, Adam Kapelner, Justin Bleich, and Emil Pitkin.
\newblock Peeking inside the black box: Visualizing statistical learning with
  plots of individual conditional expectation.
\newblock {\em Journal of Computational and Graphical Statistics},
  24(1):44--65, 2015.

\bibitem{goodrich2001experiments}
Michael~A Goodrich, Dan~R Olsen, Jacob~W Crandall, and Thomas~J Palmer.
\newblock Experiments in adjustable autonomy.
\newblock In {\em Proceedings of IJCAI Workshop on Autonomy, Delegation and
  Control: Interacting with Intelligent Agents}, pages 1624--1629. Seattle, WA:
  American Association for Artificial Intelligence Press, 2001.

\bibitem{gregor1999explanations}
Shirley Gregor and Izak Benbasat.
\newblock Explanations from intelligent systems: Theoretical foundations and
  implications for practice.
\newblock {\em MIS quarterly}, pages 497--530, 1999.

\bibitem{grudin1989case}
Jonathan Grudin.
\newblock The case against user interface consistency.
\newblock {\em Communications of the ACM}, 32(10):1164--1173, 1989.

\bibitem{Guidotti2018}
Riccardo Guidotti, Anna Monreale, Salvatore Ruggieri, Franco Turini, Fosca
  Giannotti, and Dino Pedreschi.
\newblock A survey of methods for explaining black box models.
\newblock {\em ACM Comput. Surv.}, 51(5):93:1--93:42, August 2018.

\bibitem{gunning2017explainable}
David Gunning.
\newblock Explainable artificial intelligence (xai).
\newblock {\em Defense Advanced Research Projects Agency (DARPA)}, 2017.

\bibitem{guo2010novel}
Chenlei Guo and Liming Zhang.
\newblock A novel multiresolution spatiotemporal saliency detection model and
  its applications in image and video compression.
\newblock {\em IEEE Trans. Image Processing}, 19(1):185--198, 2010.

\bibitem{guyon2003introduction}
Isabelle Guyon and Andr{\'e} Elisseeff.
\newblock An introduction to variable and feature selection.
\newblock {\em Journal of machine learning research}, 3:1157--1182, 2003.

\bibitem{hall1999correlation}
Mark~A. Hall.
\newblock Correlation-based feature selection for machine learning.
\newblock Technical report, The University of Waikato, 1999.

\bibitem{hara2016making}
Satoshi Hara and Kohei Hayashi.
\newblock Making tree ensembles interpretable.
\newblock {\em arXiv preprint arXiv:1606.05390}, 2016.

\bibitem{hart1988development}
Sandra~G Hart and Lowell~E Staveland.
\newblock Development of nasa-tlx (task load index): Results of empirical and
  theoretical research.
\newblock In {\em Advances in psychology}, volume~52, pages 139--183. Elsevier,
  1988.

\bibitem{HendricksARDSD16}
Lisa~Anne Hendricks, Zeynep Akata, Marcus Rohrbach, Jeff Donahue, Bernt
  Schiele, and Trevor Darrell.
\newblock Generating visual explanations.
\newblock In {\em Proceedings of the European Conference on Computer Vision
  (ECCV)}, pages 3--19, 2016.

\bibitem{hoffman2017explaining}
Robert~R Hoffman and Gary Klein.
\newblock Explaining explanation, part 1: theoretical foundations.
\newblock {\em IEEE Intelligent Systems}, (3):68--73, 2017.

\bibitem{holzinger2017we}
Andreas Holzinger, Chris Biemann, Constantinos~S Pattichis, and Douglas~B Kell.
\newblock What do we need to build explainable ai systems for the medical
  domain?
\newblock {\em arXiv preprint arXiv:1712.09923}, 2017.

\bibitem{hooker2004discovering}
Giles Hooker.
\newblock Discovering additive structure in black box functions.
\newblock In {\em Proceedings of the tenth ACM SIGKDD international conference
  on Knowledge discovery and data mining}, pages 575--580, 2004.

\bibitem{hu2018explainable}
Ronghang Hu, Jacob Andreas, Trevor Darrell, and Kate Saenko.
\newblock Explainable neural computation via stack neural module networks.
\newblock In {\em Proceedings of the European Conference on Computer Vision
  (ECCV)}, pages 53--69, 2018.

\bibitem{jennings2014human}
Nicholas~R Jennings, Luc Moreau, David Nicholson, Sarvapali Ramchurn, Stephen
  Roberts, Tom Rodden, and Alex Rogers.
\newblock Human-agent collectives.
\newblock {\em Communications of the ACM}, 57(12):80--88, 2014.

\bibitem{4938655}
U.~Johansson and L.~Niklasson.
\newblock Evolving decision trees using oracle guides.
\newblock In {\em 2009 IEEE Symposium on Computational Intelligence and Data
  Mining}, pages 238--244, March 2009.

\bibitem{Kahramanli}
Humar Kahramanli and Novruz Allahverdi.
\newblock Rule extraction from trained adaptive neural networks using
  artificial immune systems.
\newblock {\em Expert Syst. Appl.}, 36(2):1513--1522, March 2009.

\bibitem{katafigiotis2018stone}
Ioannis Katafigiotis, Itay~M Sabler, Eliyahu~M Heifetz, Avi Rosenfeld, Stavros
  Sfoungaristos, Amitay Lorber, Arie Latke, Vladimir Yutkin, Guy Hidas,
  Ezekiel~H Landau, et~al.
\newblock “stone-less” or negative ureteroscopy. a reality in the
  endourologic routine or avoidable source of frustration? estimating the risk
  factors for a negative ureteroscopy.
\newblock {\em Journal of endourology}, (To appear), 2018.

\bibitem{kim2016examples}
Been Kim, Rajiv Khanna, and Oluwasanmi~O Koyejo.
\newblock Examples are not enough, learn to criticize! criticism for
  interpretability.
\newblock In {\em Advances in Neural Information Processing Systems}, pages
  2280--2288, 2016.

\bibitem{kim2014bayesian}
Been Kim, Cynthia Rudin, and Julie~A Shah.
\newblock The bayesian case model: A generative approach for case-based
  reasoning and prototype classification.
\newblock In {\em Advances in Neural Information Processing Systems}, pages
  1952--1960, 2014.

\bibitem{kim2018textual}
Jinkyu Kim, Anna Rohrbach, Trevor Darrell, John Canny, and Zeynep Akata.
\newblock Textual explanations for self-driving vehicles.
\newblock In {\em Proceedings of the European Conference on Computer Vision
  (ECCV)}, pages 563--578, 2018.

\bibitem{kleinerman2018providing}
Akiva Kleinerman, Ariel Rosenfeld, and Sarit Kraus.
\newblock Providing explanations for recommendations in reciprocal
  environments.
\newblock In {\em Proceedings of the 12th ACM Conference on Recommender
  Systems}, pages 22--30. ACM, 2018.

\bibitem{knijnenburg2012explaining}
Bart~P Knijnenburg, Martijn~C Willemsen, Zeno Gantner, Hakan Soncu, and Chris
  Newell.
\newblock Explaining the user experience of recommender systems.
\newblock {\em User Modeling and User-Adapted Interaction}, 22(4-5):441--504,
  2012.

\bibitem{kofod2008explanatory}
Anders Kofod-Petersen, J{\"o}rg Cassens, and Agnar Aamodt.
\newblock Explanatory capabilities in the creek knowledge-intensive case-based
  reasoner.
\newblock {\em FRONTIERS IN ARTIFICIAL INTELLIGENCE AND APPLICATIONS}, 173:28,
  2008.

\bibitem{kononenko1993inductive}
Igor Kononenko.
\newblock Inductive and bayesian learning in medical diagnosis.
\newblock {\em Applied Artificial Intelligence an International Journal},
  7(4):317--337, 1993.

\bibitem{Kononenko99}
Igor Kononenko.
\newblock Explaining classifications for individual instances.
\newblock In {\em In Proceedings of IJCAI’99}, pages 722--726, 1999.

\bibitem{krause2016interacting}
Josua Krause, Adam Perer, and Kenney Ng.
\newblock Interacting with predictions: Visual inspection of black-box machine
  learning models.
\newblock In {\em Proceedings of the 2016 CHI Conference on Human Factors in
  Computing Systems}, pages 5686--5697, 2016.

\bibitem{KRISHNAN1999}
R.~Krishnan, G.~Sivakumar, and P.~Bhattacharya.
\newblock Extracting decision trees from trained neural networks.
\newblock {\em Pattern Recognition}, 32(12):1999 -- 2009, 1999.

\bibitem{kwon2004applying}
Oh~Byung Kwon and Norman Sadeh.
\newblock Applying case-based reasoning and multi-agent intelligent system to
  context-aware comparative shopping.
\newblock {\em Decision Support Systems}, 37(2):199--213, 2004.

\bibitem{langley2017explainable}
Pat Langley, Ben Meadows, Mohan Sridharan, and Dongkyu Choi.
\newblock Explainable agency for intelligent autonomous systems.
\newblock In {\em AAAI}, pages 4762--4764, 2017.

\bibitem{last2004compact}
Mark Last and Oded Maimon.
\newblock A compact and accurate model for classification.
\newblock {\em IEEE Transactions on Knowledge and Data Engineering},
  16(2):203--215, 2004.

\bibitem{lavravc1999selected}
Nada Lavra{\v{c}}.
\newblock Selected techniques for data mining in medicine.
\newblock {\em Artificial intelligence in medicine}, 16(1):3--23, 1999.

\bibitem{lee2004trust}
John~D Lee and Katrina~A See.
\newblock Trust in automation: Designing for appropriate reliance.
\newblock {\em Human factors}, 46(1):50--80, 2004.

\bibitem{lei2016rationalizing}
Tao Lei, Regina Barzilay, and Tommi Jaakkola.
\newblock Rationalizing neural predictions.
\newblock In {\em Proceedings of the 2016 Conference on Empirical Methods in
  Natural Language Processing}, pages 107--117, 2016.

\bibitem{letham2015interpretable}
Benjamin Letham, Cynthia Rudin, Tyler~H McCormick, David Madigan, et~al.
\newblock Interpretable classifiers using rules and bayesian analysis: Building
  a better stroke prediction model.
\newblock {\em The Annals of Applied Statistics}, 9(3):1350--1371, 2015.

\bibitem{inbook}
Roy Lewicki and Barbara Benedict~Bunker.
\newblock {\em Developing and Maintaining Trust in Working Relations}, pages
  114--139.
\newblock 1996.

\bibitem{Lipton16a}
Zachary~Chase Lipton.
\newblock The mythos of model interpretability.
\newblock {\em arXiv preprint arXiv:1606.05390}, 2016.

\bibitem{lombrozo2007simplicity}
Tania Lombrozo.
\newblock Simplicity and probability in causal explanation.
\newblock {\em Cognitive psychology}, 55(3):232--257, 2007.

\bibitem{LouCG12}
Yin Lou, Rich Caruana, and Johannes Gehrke.
\newblock Intelligible models for classification and regression.
\newblock In {\em The 18th {ACM} {SIGKDD} International Conference on Knowledge
  Discovery and Data Mining}, pages 150--158, 2012.

\bibitem{lou2013accurate}
Yin Lou, Rich Caruana, Johannes Gehrke, and Giles Hooker.
\newblock Accurate intelligible models with pairwise interactions.
\newblock In {\em Proceedings of the 19th ACM SIGKDD international conference
  on Knowledge discovery and data mining}, pages 623--631, 2013.

\bibitem{lundberg2017unified}
Scott~M Lundberg and Su-In Lee.
\newblock A unified approach to interpreting model predictions.
\newblock In {\em Advances in Neural Information Processing Systems}, pages
  4765--4774, 2017.

\bibitem{michalski1999learning}
Ryszard~S Michalski and Kenneth~A Kaufman.
\newblock Learning patterns in noisy data: the aq approach.
\newblock In {\em Advanced Course on Artificial Intelligence}, pages 22--38.
  Springer, 1999.

\bibitem{miller2017explanation}
Tim Miller.
\newblock Explanation in artificial intelligence: insights from the social
  sciences.
\newblock {\em arXiv preprint arXiv:1706.07269}, 2017.

\bibitem{Mohamed2011}
Marghny~H. Mohamed.
\newblock Rules extraction from constructively trained neural networks based on
  genetic algorithms.
\newblock {\em Neurocomput.}, 74(17):3180--3192, October 2011.

\bibitem{mohseni2018human}
Sina Mohseni and Eric~D Ragan.
\newblock A human-grounded evaluation benchmark for local explanations of
  machine learning.
\newblock {\em arXiv preprint arXiv:1801.05075}, 2018.

\bibitem{post}
Gr{\'e}goire Montavon, Wojciech Samek, and Klaus Muller.
\newblock Methods for interpreting and understanding deep neural networks.
\newblock {\em Digital Signal Processing: A Review Journal}, 73:1--15, 2 2018.

\bibitem{mood2010logistic}
Carina Mood.
\newblock Logistic regression: Why we cannot do what we think we can do, and
  what we can do about it.
\newblock {\em European sociological review}, 26(1):67--82, 2010.

\bibitem{murphy1993exploring}
Patrick~M Murphy and Michael~J Pazzani.
\newblock Exploring the decision forest: An empirical investigation of occam's
  razor in decision tree induction.
\newblock {\em Journal of Artificial Intelligence Research}, 1:257--275, 1993.

\bibitem{ortony1987surprisingness}
Andrew Ortony and Derek Partridge.
\newblock Surprisingness and expectation failure: what's the difference?
\newblock In {\em IJCAI}, pages 106--108, 1987.

\bibitem{quinlan1986induction}
J.~Ross Quinlan.
\newblock Induction of decision trees.
\newblock {\em Machine learning}, 1(1):81--106, 1986.

\bibitem{rahwan2003towards}
Iyad Rahwan, Liz Sonenberg, and Frank Dignum.
\newblock Towards interest-based negotiation.
\newblock In {\em Proceedings of the second international joint conference on
  Autonomous agents and multiagent systems}, pages 773--780, 2003.

\bibitem{Ribeiro2016}
Marco~Tulio Ribeiro, Sameer Singh, and Carlos Guestrin.
\newblock Why should {I} trust you?: Explaining the predictions of any
  classifier.
\newblock In {\em Proceedings of the 22Nd ACM SIGKDD International Conference
  on Knowledge Discovery and Data Mining}, pages 1135--1144, 2016.

\bibitem{richardson2008coach}
Ariella Richardson, Sarit Kraus, Patrice~L Weiss, and Sara Rosenblum.
\newblock Coach-cumulative online algorithm for classification of handwriting
  deficiencies.
\newblock In {\em AAAI}, pages 1725--1730, 2008.

\bibitem{rosenfeld2017intelligent}
Ariel Rosenfeld, Noa Agmon, Oleg Maksimov, and Sarit Kraus.
\newblock Intelligent agent supporting human--multi-robot team collaboration.
\newblock {\em Artificial Intelligence}, 252:211--231, 2017.

\bibitem{rosenfeld2016providing}
Ariel Rosenfeld and Sarit Kraus.
\newblock Strategical argumentative agent for human persuasion.
\newblock In {\em ECAI}, volume~16, pages 320--329, 2016.

\bibitem{rosenfeld2012learning}
Avi Rosenfeld, Zevi Bareket, Claudia~V Goldman, Sarit Kraus, David~J LeBlanc,
  and Omer Tsimhoni.
\newblock Learning driver's behavior to improve the acceptance of adaptive
  cruise control.
\newblock In {\em IAAI}, 2012.

\bibitem{rosenfeld2015learning}
Avi Rosenfeld, Zevi Bareket, Claudia~V Goldman, David~J LeBlanc, and Omer
  Tsimhoni.
\newblock Learning drivers’ behavior to improve adaptive cruise control.
\newblock {\em Journal of Intelligent Transportation Systems}, 19(1):18--31,
  2015.

\bibitem{RosenfeldSGBHL14}
Avi Rosenfeld, Vinay Sehgal, David~G. Graham, Matthew~R. Banks, Rehan~J.
  Haidry, and Laurence~B. Lovat.
\newblock Using data mining to help detect dysplasia: Extended abstract.
\newblock In {\em 2014 {IEEE} International Conference on Software Science,
  Technology and Engineering,}, pages 65--66, 2014.

\bibitem{rosenfeld2016negochat}
Avi Rosenfeld, Inon Zuckerman, Erel Segal-Halevi, Osnat Drein, and Sarit Kraus.
\newblock Negochat-a: a chat-based negotiation agent with bounded rationality.
\newblock {\em Autonomous Agents and Multi-Agent Systems}, 30(1):60--81, 2016.

\bibitem{rubio2004evaluation}
Susana Rubio, Eva D{\'\i}az, Jes{\'u}s Mart{\'\i}n, and Jos{\'e}~M Puente.
\newblock Evaluation of subjective mental workload: A comparison of swat,
  nasa-tlx, and workload profile methods.
\newblock {\em Applied Psychology}, 53(1):61--86, 2004.

\bibitem{rudin2014algorithms}
Cynthia Rudin.
\newblock Algorithms for interpretable machine learning.
\newblock In {\em Proceedings of the 20th ACM SIGKDD international conference
  on Knowledge discovery and data mining}, pages 1519--1519, 2014.

\bibitem{Saeys2007survey}
Yvan Saeys, Inaki Inza, and Pedro Larra\~{n}aga.
\newblock A review of feature selection techniques in bioinformatics.
\newblock {\em Bioinformatics}, 23(19):2507--2517, 2007.

\bibitem{salem2015would}
Maha Salem, Gabriella Lakatos, Farshid Amirabdollahian, and Kerstin Dautenhahn.
\newblock Would you trust a (faulty) robot?: Effects of error, task type and
  personality on human-robot cooperation and trust.
\newblock In {\em Proceedings of the Tenth Annual ACM/IEEE International
  Conference on Human-Robot Interaction}, pages 141--148, 2015.

\bibitem{saltelli2002sensitivity}
Andrea Saltelli.
\newblock Sensitivity analysis for importance assessment.
\newblock {\em Risk analysis}, 22(3):579--590, 2002.

\bibitem{samek2017explainable}
Wojciech Samek, Thomas Wiegand, and Klaus-Robert M{\"u}ller.
\newblock Explainable artificial intelligence: Understanding, visualizing and
  interpreting deep learning models.
\newblock {\em arXiv preprint arXiv:1708.08296}, 2017.

\bibitem{scerri2001adjustable}
Paul Scerri, David Pynadath, and Milind Tambe.
\newblock Adjustable autonomy in real-world multi-agent environments.
\newblock In {\em Proceedings of the fifth international conference on
  Autonomous agents}, pages 300--307. ACM, 2001.

\bibitem{schank1986explanation}
Roger~C Schank.
\newblock Explanation: A first pass.
\newblock {\em Experience, memory, and reasoning}, pages 139--165, 1986.

\bibitem{4167900}
V.~Schetinin, J.~E. Fieldsend, D.~Partridge, T.~J. Coats, W.~J. Krzanowski,
  R.~M. Everson, T.~C. Bailey, and A.~Hernandez.
\newblock Confident interpretation of bayesian decision tree ensembles for
  clinical applications.
\newblock {\em IEEE Transactions on Information Technology in Biomedicine},
  11(3):312--319, May 2007.

\bibitem{selvaraju2017grad}
Ramprasaath~R Selvaraju, Michael Cogswell, Abhishek Das, Ramakrishna Vedantam,
  Devi Parikh, Dhruv Batra, et~al.
\newblock Grad-cam: Visual explanations from deep networks via gradient-based
  localization.
\newblock In {\em ICCV}, pages 618--626, 2017.

\bibitem{sheh2017did}
Raymond Sheh.
\newblock why did you do that?” explainable intelligent robots.
\newblock In {\em AAAI Workshop on Human-Aware Artificial Intelligence}, 2017.

\bibitem{shneiderman2002promoting}
Ben Shneiderman.
\newblock Promoting universal usability with multi-layer interface design.
\newblock {\em ACM SIGCAPH Computers and the Physically Handicapped},
  (73-74):1--8, 2002.

\bibitem{shrot2014crisp}
Tammar Shrot, Avi Rosenfeld, Jennifer Golbeck, and Sarit Kraus.
\newblock Crisp: an interruption management algorithm based on collaborative
  filtering.
\newblock In {\em Proceedings of the SIGCHI conference on human factors in
  computing systems}, pages 3035--3044, 2014.

\bibitem{shwartz2017opening}
Ravid Shwartz-Ziv and Naftali Tishby.
\newblock Opening the black box of deep neural networks via information.
\newblock {\em arXiv preprint arXiv:1703.00810}, 2017.

\bibitem{sierhuis2003human}
Maarten Sierhuis, Jeffrey~M Bradshaw, Alessandro Acquisti, Ron Van~Hoof, Renia
  Jeffers, and Andrzej Uszok.
\newblock Human-agent teamwork and adjustable autonomy in practice.
\newblock In {\em Proceedings of the seventh international symposium on
  artificial intelligence, robotics and automation in space (I-SAIRAS)}, 2003.

\bibitem{Simonyan2013DeepIC}
Karen Simonyan, Andrea Vedaldi, and Andrew Zisserman.
\newblock Deep inside convolutional networks: Visualising image classification
  models and saliency maps.
\newblock {\em CoRR}, abs/1312.6034, 2013.

\bibitem{sormo2004explanation}
Frode S{\o}rmo and J{\"o}rg Cassens.
\newblock Explanation goals in case-based reasoning.
\newblock In {\em Proceedings of the ECCBR 2004 Workshops}, number 142-04,
  pages 165--174, 2004.

\bibitem{sormo2005explanation}
Frode S{\o}rmo, J{\"o}rg Cassens, and Agnar Aamodt.
\newblock Explanation in case-based reasoning--perspectives and goals.
\newblock {\em Artificial Intelligence Review}, 24(2):109--143, 2005.

\bibitem{SteinGNRJ17}
Sebastian Stein, Enrico~H. Gerding, Adrian Nedea, Avi Rosenfeld, and
  Nicholas~R. Jennings.
\newblock Market interfaces for electric vehicle charging.
\newblock {\em J. Artif. Intell. Res.}, 59:175--227, 2017.

\bibitem{Strumbelj2010}
Erik Strumbelj and Igor Kononenko.
\newblock An efficient explanation of individual classifications using game
  theory.
\newblock {\em J. Mach. Learn. Res.}, 11:1--18, March 2010.

\bibitem{tan2016tree}
Hui~Fen Tan, Giles Hooker, and Martin~T Wells.
\newblock Tree space prototypes: Another look at making tree ensembles
  interpretable.
\newblock {\em arXiv preprint arXiv:1611.07115}, 2016.

\bibitem{tolomei2017interpretable}
Gabriele Tolomei, Fabrizio Silvestri, Andrew Haines, and Mounia Lalmas.
\newblock Interpretable predictions of tree-based ensembles via actionable
  feature tweaking.
\newblock In {\em Proceedings of the 23rd ACM SIGKDD International Conference
  on Knowledge Discovery and Data Mining}, pages 465--474, 2017.

\bibitem{traum2003negotiation}
David Traum, Jeff Rickel, Jonathan Gratch, and Stacy Marsella.
\newblock Negotiation over tasks in hybrid human-agent teams for
  simulation-based training.
\newblock In {\em Proceedings of the second international joint conference on
  Autonomous agents and multiagent systems}, pages 441--448. ACM, 2003.

\bibitem{van198511}
Bas~C Van~Fraassen.
\newblock 11 empiricism in the philosophy of science.
\newblock {\em Images of science: Essays on realism and empiricism, with a
  reply from Bas C. van Fraassen}, page 245, 1985.

\bibitem{vanlehn2011relative}
Kurt VanLehn.
\newblock The relative effectiveness of human tutoring, intelligent tutoring
  systems, and other tutoring systems.
\newblock {\em Educational Psychologist}, 46(4):197--221, 2011.

\bibitem{vellido2012robust}
A~Vellido, E~Romero, M~Juli{\`a}-Sap{\'e}, C~Maj{\'o}s, {\`A}~Moreno-Torres,
  J~Pujol, and C~Ar{\'u}s.
\newblock Robust discrimination of glioblastomas from metastatic brain tumors
  on the basis of single-voxel 1h mrs.
\newblock {\em NMR in Biomedicine}, 25(6):819--828, 2012.

\bibitem{vellido2012making}
Alfredo Vellido, Jos{\'e}~David Mart{\'\i}n-Guerrero, and Paulo~JG Lisboa.
\newblock Making machine learning models interpretable.
\newblock In {\em ESANN}, volume~12, pages 163--172, 2012.

\bibitem{vigano2018explainable}
Luca Vigan{\`o} and Daniele Magazzeni.
\newblock Explainable security.
\newblock {\em arXiv preprint arXiv:1807.04178}, 2018.

\bibitem{vlek2016method}
Charlotte~S Vlek, Henry Prakken, Silja Renooij, and Bart Verheij.
\newblock A method for explaining bayesian networks for legal evidence with
  scenarios.
\newblock {\em Artificial Intelligence and Law}, 24(3):285--324, 2016.

\bibitem{wang2017bayesian}
Tong Wang, Cynthia Rudin, Finale Doshi-Velez, Yimin Liu, Erica Klampfl, and
  Perry MacNeille.
\newblock A bayesian framework for learning rule sets for interpretable
  classification.
\newblock {\em The Journal of Machine Learning Research}, 18(1):2357--2393,
  2017.

\bibitem{whitmore2018explicating}
Leanne~S Whitmore, Anthe George, and Corey~M Hudson.
\newblock Explicating feature contribution using random forest proximity
  distances.
\newblock {\em arXiv preprint arXiv:1807.06572}, 2018.

\bibitem{xiao2007commerce}
Bo~Xiao and Izak Benbasat.
\newblock E-commerce product recommendation agents: use, characteristics, and
  impact.
\newblock {\em MIS quarterly}, 31(1):137--209, 2007.

\bibitem{xu2015show}
Kelvin Xu, Jimmy Ba, Ryan Kiros, Kyunghyun Cho, Aaron Courville, Ruslan
  Salakhudinov, Rich Zemel, and Yoshua Bengio.
\newblock Show, attend and tell: Neural image caption generation with visual
  attention.
\newblock In {\em International conference on machine learning}, pages
  2048--2057, 2015.

\bibitem{yanco2004classifying}
Holly~A Yanco and Jill Drury.
\newblock Classifying human-robot interaction: an updated taxonomy.
\newblock In {\em Systems, Man and Cybernetics, 2004 IEEE International
  Conference on}, volume~3, pages 2841--2846. IEEE, 2004.

\bibitem{yetim2008framework}
Fahri Yetim.
\newblock A framework for organizing justifications for strategic use in
  adaptive interaction contexts.
\newblock In {\em ECIS}, pages 815--825, 2008.

\bibitem{yosinski2015understanding}
Jason Yosinski, Jeff Clune, Anh Nguyen, Thomas Fuchs, and Hod Lipson.
\newblock Understanding neural networks through deep visualization.
\newblock {\em arXiv preprint arXiv:1506.06579}, 2015.

\bibitem{zhang2018visual}
Quan-shi Zhang and Song-Chun Zhu.
\newblock Visual interpretability for deep learning: a survey.
\newblock {\em Frontiers of Information Technology \& Electronic Engineering},
  19(1):27--39, 2018.

\bibitem{zhang2015sensitivity}
Ye~Zhang and Byron Wallace.
\newblock A sensitivity analysis of (and practitioners' guide to) convolutional
  neural networks for sentence classification.
\newblock {\em arXiv preprint arXiv:1510.03820}, 2015.

\bibitem{zhou2016learning}
Bolei Zhou, Aditya Khosla, Agata Lapedriza, Aude Oliva, and Antonio Torralba.
\newblock Learning deep features for discriminative localization.
\newblock In {\em Proceedings of the IEEE Conference on Computer Vision and
  Pattern Recognition}, pages 2921--2929, 2016.

\bibitem{zhou2016interpreting}
Yichen Zhou and Giles Hooker.
\newblock Interpreting models via single tree approximation.
\newblock {\em arXiv preprint arXiv:1610.09036}, 2016.

\bibitem{zhou2003extracting}
Zhi-Hua Zhou, Yuan Jiang, and Shi-Fu Chen.
\newblock Extracting symbolic rules from trained neural network ensembles.
\newblock {\em Ai Communications}, 16(1):3--15, 2003.

\end{thebibliography}

\end{document}